\documentclass[journal,twoside,web]{ieeecolor}
\usepackage{tmi}
\usepackage{cite}
\usepackage{amsmath,amssymb,amsfonts}
\usepackage{textcomp}
\usepackage[pdftex]{graphicx}
\usepackage[misc]{ifsym}
\usepackage[linesnumbered,ruled]{algorithm2e}
\usepackage{color, colortbl}
\usepackage{spverbatim}
\usepackage{etoolbox}
\usepackage{multirow}
\usepackage{booktabs}
\usepackage{courier}
\usepackage{bm}
\usepackage{bbm}
\usepackage{epstopdf}

\def\BibTeX{{\rm B\kern-.05em{\sc i\kern-.025em b}\kern-.08em
    T\kern-.1667em\lower.7ex\hbox{E}\kern-.125emX}}
\begin{document}
\title{Learning Inductive Attention Guidance for \\ Partially Supervised Pancreatic Ductal Adenocarcinoma Prediction}
\author{Yan~Wang,
        Peng~Tang,
        Yuyin~Zhou,
        Wei~Shen,
        Elliot~K.~Fishman,
        and~Alan~L.~Yuille

\thanks{Manuscript submitted on Jan 7, 2021. This work was supported by the Lustgarten Foundation for Pancreatic Cancer Research.}
\thanks{Y. Wang is with Shanghai Key Laboratory of Multidimensional Information Processing, East China Normal University, Shanghai, China, and also with Department of Computer Science, the Johns Hopkins University Baltimore, MD 21218, USA (e-mail: wyanny.9@gmail.com).}
\thanks{P. Tang, Y. Zhou, and A. L. Yuille are with the Department of Computer Science,
    the Johns Hopkins University Baltimore, MD 21218, USA
    (e-mail: \{tangpeng723, zhouyuyiner, alan.l.yuille\}@gmail.com).}
\thanks{W. Shen is with MoE Key Lab of Artificial Intelligence, AI Institute, Shanghai Jiao Tong University, Shanghai, China, and also with Department of Computer Science, the Johns Hopkins University Baltimore, MD 21218, USA (e-mail: shenwei1231@gmail.com).}
\thanks{E. K. Fishman is with the Department of Radiology,
    the Johns Hopkins University School of Medicine, Baltimore, MD 21287, USA
    (e-mail: efishman@jhmi.edu). }
\thanks{Corresponding author: W. Shen}}

\maketitle

\begin{abstract}
Pancreatic ductal adenocarcinoma (PDAC) is the third most common cause of cancer death in the United States. Predicting tumors like PDACs (including both classification and segmentation) from medical images by deep learning is becoming a growing trend, but usually a large number of annotated data are required for training, which is very labor-intensive and time-consuming. In this paper, we consider a partially supervised setting, where cheap image-level annotations are provided for all the training data, and the costly per-voxel annotations are only available for a subset of them. We propose an Inductive Attention Guidance Network (IAG-Net) to jointly learn a global image-level classifier for normal/PDAC classification and a local voxel-level classifier for semi-supervised PDAC segmentation. We instantiate both the global and the local classifiers by multiple instance learning (MIL), where the attention guidance, indicating roughly where the PDAC regions are, is the key to bridging them: For global MIL based normal/PDAC classification, attention serves as a weight for each instance (voxel) during MIL pooling, which eliminates the distraction from the background; For local MIL based semi-supervised PDAC segmentation, the attention guidance is inductive, which not only provides bag-level pseudo-labels to training data without per-voxel annotations for MIL training, but also acts as a proxy of an instance-level classifier. Experimental results show that our IAG-Net boosts PDAC segmentation accuracy by {more than 5\%} compared with the state-of-the-arts.
\end{abstract}


\section{Introduction}
\label{sec:introduction}
\IEEEPARstart{P}ancreatic ductal adenocarcinoma (PDAC) is one of the most deadly diseases, whose prognosis is dismal as more than 50\% of patients have evidence of metastatic disease at the time of diagnosis. Currently, detecting or segmenting PDACs through medical imaging at the localized disease stage followed by complete resection can offer the best chance of survival \cite{Zhou2019hyper}. Computed tomography (CT) screening is the most commonly used imaging modality for the initial evaluation of PDACs. However, finding PDACs in CT images is challenging, even for experienced radiologists. Therefore, to build up computer-aided diagnosis (CAD) systems with the ability to automatically identify suspicious cases and alert radiologists is vital.
In clinical environments, this cannot be simply formulated as either a classification or localization/segmentation problem, but a joint problem of these two tasks. Our goal is to address the PDAC cancer prediction problem: given a CT scan of a patient, we need to determine (\emph{i.e.}, classify) whether this patient suffers from PDAC cancer or not, and if yes, to localize where the PDAC region is. The latter is of great importance as it provides a visual interpretation to support the former result.

\begin{figure}[t]
\begin{center}
    \includegraphics[width=0.9\linewidth]{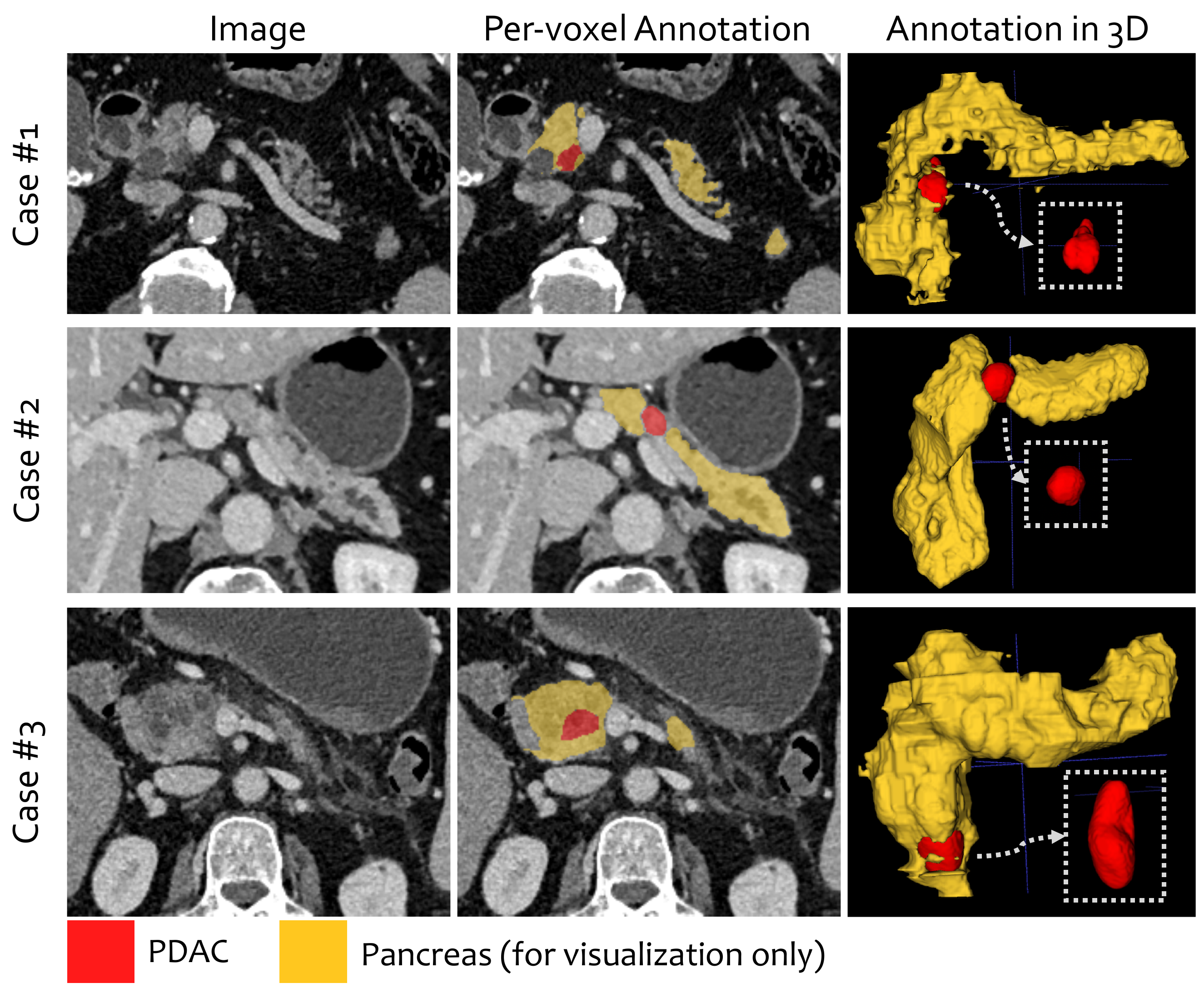}
\end{center}
\vspace{-0.5em}
\caption{PDAC examples shown in 2D and rendered in 3D. Per-voxel pancreas labels are also shown as reference. Since part of the PDAC can be inside the pancreas, the whole PDAC is shown in white dashed boxes in 3D.} 
\label{Fig:Dataset}
\vspace{-1.5em}
\end{figure}

A growing literature has proposed various techniques based on supervised learning for medical image segmentation/localization \cite{roth2015deeporgan,roth2016spatial,Zhou2020unet,Imran2019fast,Yan2018deeplesion,dou20163D}. But these methods require a large amount of per-voxel annotated data. Usually, this high-quality manual contouring process is not only tedious, but also expensive (considering the cost of salaries, segmentation software, and training). This situation is more conspicuous for PDAC segmentation in CT images as PDAC masses in the early stage are usually small and their boundaries are usually weak, which can even confuse radiologists. Thus, we consider addressing the PDAC cancer prediction problem under a (weakly) partially supervised setting: we are given the training data with image-level annotations, \emph{i.e.}, we know whether each CT image has PDAC masses or not, but only a small subset of them have per-voxel annotations, \emph{i.e.}, we know whether each voxel belongs to a PDAC region or not.

Under this partially supervised setting, to address the PDAC cancer prediction problem, \emph{i.e.}, the joint (multi-task) problem of normal/PDAC classification and PDAC segmentation, two types of weakly supervised learning techniques can be used. One is inexact supervised learning (ISL)~\cite{Zhou2018a}, such as multiple instance learning (MIL), the other is semi-supervised learning (SSL)~\cite{Zhuphdthesis}, such as Expectation-Maximization (EM) \cite{Yuille2018Deep}. MIL based methods build a bag-level (image-level) classifier upon bag representations aggregated from instance (voxel) features over the whole image{,} with the ability to infer per-voxel labels from image-level annotations~\cite{Xu2012multiple,Yao2018weakly,Couture18multiple}. EM-like methods make use of the large amount of training data without per-voxel annotations in conjunction with the small amount of training data with per-voxel annotations to improve instance-level classifiers for localization/segmentation~\cite{Bai2017semi,Bortsova2019semi,Wang2020focalmix,Zhou2019prior}. 
Nevertheless, due to the difficulties in PDAC segmentation in the early stage, such as small sizes, weak boundaries, variety of shapes and diverse locations with a pancreas region (see Fig.~\ref{Fig:Dataset}), current MIL-based classification methods or EM-like segmentation methods or multi-task learning methods cannot achieve satisfactory results.
The MIL-based methods, \emph{e.g.}, \cite{Li2018thoracic}, used a classic MIL operator ({\em{i.e.},} xor) to aggregate all the instance features to form the bag representation, which is vulnerable to the distraction from {the} background, since a PDAC is usually much smaller than the whole pancreas region; The EM-like methods, \emph{e.g.}, \cite{Zhou2019semi} relied on self-training strategies~\cite{Ref:Scudder65a}. But the generated per-voxel pseudo-labels are often heavily noisy due to the difficulties in PDAC segmentation; The multi-task learning methods, \emph{e.g.}, \cite{Shin2019joint}, essentially conducted these two learning tasks separately, except for shared feature learning, and thus suffer from both above issues.

\begin{figure}[t]
\begin{center}
    \includegraphics[width=1\linewidth]{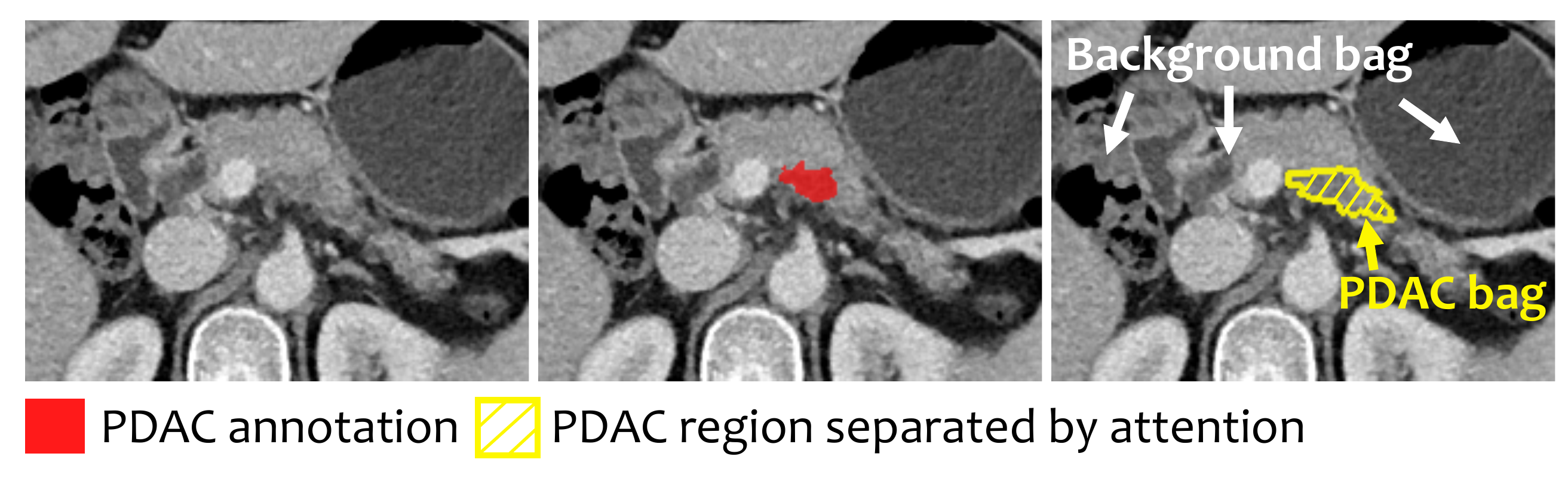}
\end{center}
\vspace{-0.7em}
\caption{The PDAC region separated by attention consists of background voxels (middle \& right images). Per-voxel pseudo-label is noisy, but from the MIL perspective, the bag-level pseudo-labels assigned according to the separation are correct.} 
\label{Fig:attention_example}
\vspace{-1.3em}
\end{figure}

To cope with the above difficulties, in this paper, we propose an attention-guided framework to jointly learn a global (image-level) classifier for normal/PDAC classification and a local (instance-level) classifier for semi-supervised PDAC segmentation. 
The attention guidance, indicating roughly where PDAC regions are, is explicitly learned from the training data with per-voxel annotations and inducted on the training data without per-voxel annotations. In this framework, both the global and the local classifiers are instantiated by MIL and the attention guidance is the key to bridging them.  For normal/PDAC classification, the attention guidance serves as a weight for each instance, resulting in weighted MIL pooling to suppress the distraction from background when training the global MIL classifier; For semi-supervised PDAC segmentation, the attention guidance {separates the PDAC and the background regions} on the training data without per-voxel annotations as bag-level pseudo labels for training the local MIL classifier in the EM manner. Since the instances of these training data without per-voxel annotations are ``bagged'' instead of treated as singletons, the local MIL classifier offers a possible way to mitigating the effects of noises in per-voxel pseudo labels~\cite{Ref:LeungSZ11}, as shown in Fig.~\ref{Fig:attention_example}.  Note that, the attention guidance is {\bf{inductive}}~\cite{Ref:SarkarH06}, since it can not only provide bag-level pseudo-labels to training data without per-voxel annotations, but also act as a proxy of a local instance-level classifier defined on all the data which have PDAC masses. 

We instantiate our attention-guided framework by a single network with two streams. The backbone of the network, \emph{e.g.}, a VGG-Net~\cite{simonyan2015very} or an U-Net~\cite{Ronneberger2015u}, generates a convolutional feature map from an input image, in which each feature vector at a spatial location is an instance. Then the attention is learned explicitly on the feature map from training data with per-voxel annotations. 
The feature map is further branched into two streams: The first stream trains the global MIL classifier by using the attention guidance {as a weighting mechanism for instances} in MIL pooling; The second stream first separates PDAC regions and background regions on the training data without per-voxel annotations according to the attention guidance, and then trains the local MIL classifier by bagging the instances in the PDAC regions and  background regions, respectively. We refer to this network with two streams as {\bf{IAG-Net}}, for \textbf{I}nductive \textbf{A}ttention \textbf{G}uidance \textbf{Net}work.

Given a training set consisting of normal ({\em{i.e.}}, healthy cases) and abnormal ({\em{i.e.}}, cases diagnosed with PDAC) CT scans under the partially supervised setting, all the streams and the backbone of IAG-Net can be jointly trained. It provides both image-level (\emph{i.e.}, normal/PDAC classification) results and corresponding voxel-level (\emph{i.e.}, PDAC segmentation) visual evidences. Experimental results show such a design performs favorably for PDAC prediction, boosting PDAC segmentation results by a large margin. 

The contribution of this paper is four-fold:
\begin{enumerate}
\item We propose an attention-guided framework to address classification and segmentation of PDAC under {the} partially supervised setting, which is {of great potential for practical applications.} 
\item Unlike previous semi-supervised segmentation methods, which train segmentors (instance-level classifiers) on per-voxel pseudo-labels, our framework trains the instance-level classifier by MIL. It takes the {pseudo-PDAC} and  background regions as bags, which {addresses} the issue of heavy noises in per-voxel pseudo labels.
\item Our framework has active interactions between PDAC/normal classification and PDAC segmentation, bridging by the inductive attention guidance. 
\item We instantiate our framework by a single network  with two streams, IAG-Net, in which the backbone and the two streams can be trained jointly. IAG-Net achieves {a substantial improvement of around 5\% DSC for PDAC segmentation}. 

\end{enumerate}

\section{Related Work}
\label{RelatedWork}
Weakly supervised learning is widely used in the field of computer vision \cite{zhu2009introduction,Zhou2018a,tang2017multiple,tang2018pcl,papandreou2015weakly,cinbis2016weakly,Roth2018an}. In this section, we briefly review the related works on weakly-supervised learning for medical image classification and segmentation/detection.
\vspace{-1mm}
\subsection{Semi-supervised learning}

Semi-supervised learning (SSL)~\cite{Zhuphdthesis}, is also known as incomplete supervised learning \cite{Zhou2018a}, where only a small subset of training data are labeled whereas the other data remain unlabeled. The most widely used techniques for SSL are EM-like methods such as self-training \cite{Bai2017semi} and co-training \cite{Zhou2019semi}. Other directions such as consistency-based methods \cite{Bortsova2019semi,Ref:LiYCFH20,Fotedar2020extreme} are becoming popular recently.

Self-training propagates labels from the labeled to the unlabeled data, and then using the larger, newly labeled set for training. This approach assumes that the method's high confidence predictions are correct. The expectation-maximization procedure alternates between assigning pseudo-labels to the unlabeled data given the labeled data and model parameters, and updating the model parameters given all the data \cite{Bai2017semi}\cite{Zhou2019semi}. \cite{Bai2017semi} trained and predicted on {a} single plane while the work proposed in \cite{Zhou2019semi}, termed as DMPCT, distilled consensus information from three planes of {the 3D volume of a CT scan}. DMPCT \cite{Zhou2019semi} adopted the idea from co-training \cite{Blum1998combining}, where classifiers are trained with independent sets of features, and the classifiers rely on each other for estimating the confidence of their predictions. {Many SSL methods heavily rely on the quality of per-voxel pseudo-labels, which is hard to be guaranteed when the segmentation task is challenging, \emph{e.g.}, such as PDAC segmentation. {Our method belongs to self-training based methods. But }the pseudo-label generated by our method is in bag-level (see Fig.~\ref{Fig:attention_example}), addressed by MIL \cite{dietterichLL1997solving}, and it can tolerate voxel-level errors. Another research direction is consistency-based methods~\cite{Bortsova2019semi,Ref:LiYCFH20,Liu2020semi}, which encouraged consistent segmentation/classification of the network-in-training for the same input (on both labeled and unlabeled images) under a given class of transformations. These methods are complementary with ours \cite{Tang2020proposal}.}
\vspace{-1mm}
\subsection{Inexact supervised learning}

In inexact supervised learning (ISL) \cite{Zhou2018a}, only coarse-grained labels are provided. One particular form of ISL is multiple instance learning (MIL), first proposed for drug activities prediction \cite{dietterichLL1997solving}. In MIL, a training set consists of a number of bags, each of which is assigned with a positive or negative label (or multi-class label). Each bag contains a group of instances, which are not individually labeled like traditional supervised learning setting. 
The goal of MIL is to learn instance classifiers under MIL constraints ({\em{i.e.}}, a bag should be labeled as positive if at least one of its instances is positive and labeled as negative otherwise). Many previous methods in {the} medical imaging community adopted {the} multiple instance learning (MIL) pipeline for weakly supervised abnormality classification and detection. These methods have achieved encouraging results on various tasks \cite{Xu2012multiple,Yao2018weakly,Couture18multiple}. Xu \emph{et al.} \cite{Xu2012multiple} proposed a first integrated framework for histopathology cancer image classification, segmentation and clustering. Yao \emph{et al.} \cite{Yao2018weakly} embedded MIL into deep learning frameworks for thoracic disease identification and localization. Breast tumor histology classification uses quantile aggregation to predict the class of the cropped image region \cite{Couture18multiple}. But, there is still a big gap between the results of these MIL algorithms with fully supervised ones. Besides, most MIL-based algorithms \cite{Xu2012multiple,Yao2018weakly,Couture18multiple} do not have any instance label, thus localization results are far from satisfactory. 
\vspace{-1mm}
\subsection{Partially supervised learning}

There are also some methods considering partially supervised learning, which use a large number of image-level annotated data, and a subset of pixel-level annotated data \cite{Shin2019joint,Li2018thoracic,Liu2019align,Zhou2019collaborative}. Li \emph{et al.} \cite{Li2018thoracic} proposed a unified framework for disease identification and localization of abnormalities in chest X-ray {images}.
CIA-Net~\cite{Liu2019align} exploited the highly structured property of chest X-ray images which localized diseases via a pair of aligned positive and negative samples. 
Collaborative learning~\cite{Zhou2019collaborative} jointly improves the performance of disease grading and lesion segmentation on fundus images for diabetic retinopathy. It used attention maps generated by lesion	attentive classification module as pseudo-labels for SSL.

{Different from \cite{Li2018thoracic,Liu2019align} which are tailored for detection, our IAG-Net is designed for segmentation.
Additionally, our work differs from \cite{Li2018thoracic} in that: Li's work did not leverage the localization information from the data without per-voxel annotations. Meanwhile, our differences compared with \cite{Liu2019align} are 1) our IAG-Net does not rely on aligned images, while the attention map of Liu's CIA-Net relies on a pair of aligned positive and negative images; 2) we assign pseudo-labels to the data without per-voxel annotation while CIA-Net leverages the spatial-wise attention map indicating the possible location of the disease in images without bounding box annotation.}
{The differences of our work compared with Zhou's collaborative learning method \cite{Zhou2019collaborative} are 1) for images without per-voxel annotations, our IAG-Net uses bag-level pseudo-labels, while Zhou's work uses per-voxel pseudo-labels; 2) the global classifier in IAG-Net is designed under popular MIL constraints for solving classification problems where positive instances only exist in positive bags, while Zhou's work is aiming at classifying disease severity gradings, where lesion symptoms may co-exist in different classes.}

\section{Inductive Attention Guidance Network}
Mathematically, let the {3D volume of a CT} image{\footnote{3D image will be termed as {\em{image}} for short in the rest of the paper}} denoted by $\mathbf{Z}\in\mathbb{R}^{W\times H\times L}$. The goal of PDAC prediction is to predict 1) the image-level label $\hat{\mathbbm{y}}\in\{0,1\}$ of $\mathbf{Z}$, indicating whether it contains PDAC masses ($\hat{\mathbbm{y}}=1$) or not ($\hat{\mathbbm{y}}=0$), and 2) the per-voxel label map $\hat{\mathbf{Y}}\in\{0,1\}^{W\times H\times L}$, indicating where PDAC masses are in $\mathbf{Z}$. 
Our training set $\mathcal{D}$ consists of two subsets: $\mathcal{D}=\mathcal{D}^l\bigcup\mathcal{D}^u$, where $\mathcal{D}^l = \{(\mathbf{Z}^l,\mathbf{Y}^l)\}_{l=1}^L$ and $\mathcal{D}^u = \{(\mathbf{Z}^u,\mathbbm{y}^u)\}_{u=L+1}^U$. The training data in $\mathcal{D}^l$ are given by \textbf{annotated per-voxel label maps} while those in $\mathcal{D}^u$ are only given by \textbf{annotated image-level labels}. Note that, if the per-voxel label map $\mathbf{Y}$ is known, then the image-level label $\mathbbm{y}$ is also known, but not vice versa if $\mathbbm{y}=1$.

\begin{figure*}[t]
\begin{center}
    \includegraphics[width=1\linewidth]{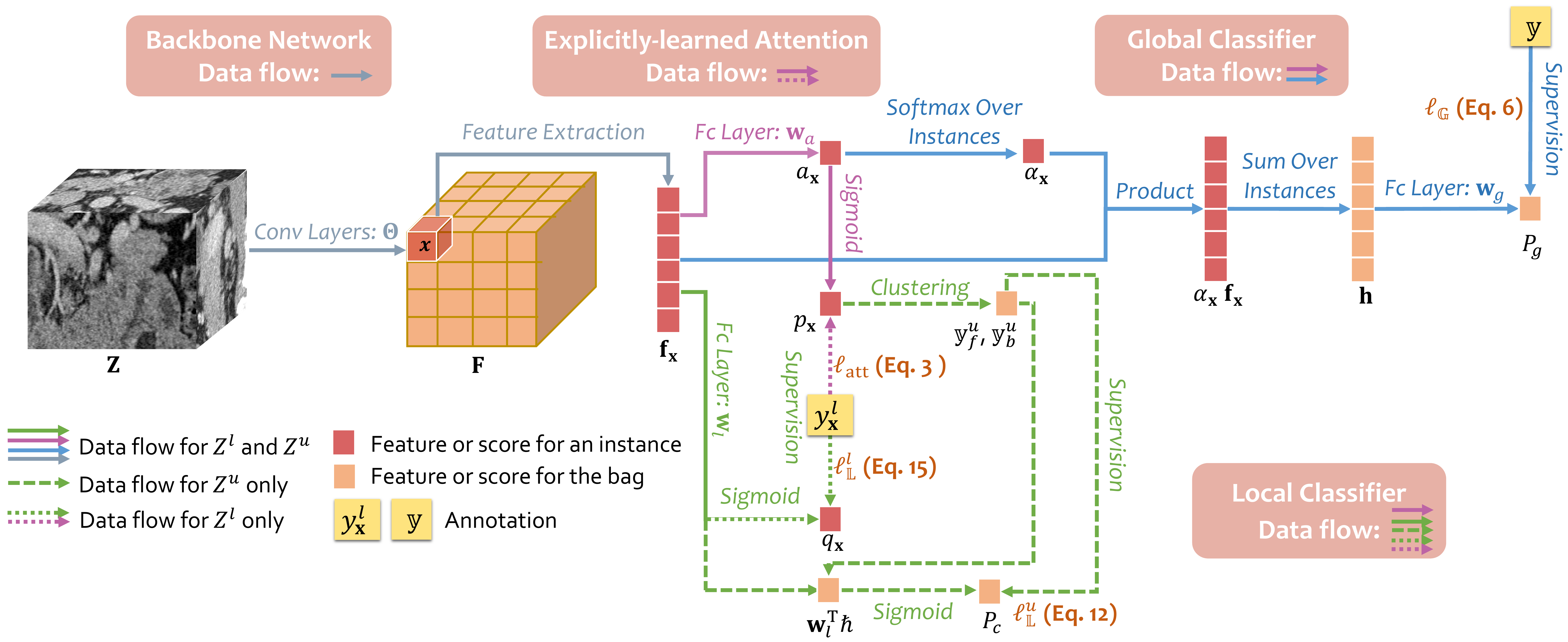}
\end{center}
\caption{{The architecture of IAG-Net.} Our IAG-Net has a backbone network and two streams: 1) Global classifier for normal/PDAC classification (Sec. \ref{sec:attention_MIL}) and 2) Local classifier (Sec. \ref{sec:local_classifier}) for PDAC segmentation. Explicitly-learned attention (Sec. \ref{Sec:el_attention}) bridges the two streams, and it is also a proxy of the local classifier. {The volume of a CT scan} $\mathbf{Z}$ is fed into a backbone network, parameterized by $\bm\Theta$. $\mathbf{F}$ is the feature map. $\mathbf{f}_\mathbf{x}$ represents the feature vector at spatial location $\mathbf{x}$ on the feature map. Attention value $a_\mathbf{x}$ is obtained through Eq.~\ref{eq:att_value}. Attention values are normalized through a softmax function to get $\alpha_\mathbf{x}$. Image-level feature $\mathbf{h}$ is obtained through Eq.~\ref{eq:bag_feature}. $P_g$ is the probability that the image belongs to PDAC (see Eq.~\ref{eq:global_prob}). If $\mathbf{Z}=\mathbf{Z}^l$, the probability-like attention value $p_\mathbf{x}$ (Eq.~\ref{eq:att_prob}) and probability $q_\mathbf{x}$ (Eq.~\ref{eq:local_prob}) are supervised by the annotation of the instance $y_\mathbf{x}^l$. If $\mathbf{Z} = \mathbf{Z}^u$, pseudo-labels $\mathbbm{y}_f^u$ and $\mathbbm{y}_b^u$ are generated by attention-based separation (Eq.~\ref{eq:loss_clustering}), which are used to train a local MIL classifier under {the} SSL setting (Eq.~\ref{eq:SSL_unlabeled}).
} 
\label{Fig:framework}
\vspace{-1em}
\end{figure*}

The overall pipeline of the proposed IAG-Net is shown in Fig.~\ref{Fig:framework}. An image $\mathbf{Z}$ is first fed into the backbone parameterized by $\bm{\Theta}$ to produce a feature map $\mathbf{F}(\mathbf{Z};\bm{\Theta})\in\mathbb{R}^{W'\times H'\times L'\times C}$, where $C$ is the number of feature channels. The feature vector $\mathbf{f}_\mathbf{x}(\mathbf{Z};\bm{\Theta})\in\mathbb{R}^C$ at each spatial location $\mathbf{x}$ on the feature map $\mathbf{F}(\mathbf{Z};\bm{\Theta})$ represents the feature of an instance. Then attention values are learned explicitly on the feature map from the training data with per-voxel annotations which have PDAC masses (Sec. \ref{Sec:el_attention}). 
The feature map is branched into two streams: The first stream performs attention-guided MIL to train a global image-level (bag-level) classifier on all the training data (Sec. \ref{sec:attention_MIL});
The second stream trains the local MIL classifier by bagging the instances in the PDAC regions and background regions inducted by attention on the training data without per-voxel annotations (Sec. \ref{sec:local_classifier}). The overall loss function for training IAG-Net is the sum of loss functions for the two streams. Next, we describe each module in detail. Finally, we show the testing procedure of IAG-Net\vspace{-0.4em}.

\subsection{Explicitly-learned Attention}
\label{Sec:el_attention}
Intuitively, high attention values highlight possible PDAC regions. Since attention acts as a proxy of an instance-level local classifier, explicitly learning attention on the training data with per-voxel annotations is straightforward. 

We define the attention value of an instance $\mathbf{f}_\mathbf{x}(\mathbf{Z};\bm{\Theta})$ as its linear projection:
\begin{equation}
a_{\mathbf{x}}(\mathbf{Z};\bm{\Theta},\mathbf{w}_a)=\mathbf{w}_a^{\top}\mathbf{f}_\mathbf{x}(\mathbf{Z};\bm{\Theta}),\label{eq:att_value}
\end{equation}
where $\mathbf{w}_a\in\mathbb{R}^C$. To learn the attention value for each instance, we first apply a sigmoid function to it to obtain a probability-like attention value:
\begin{equation}
   p_\mathbf{x}(\mathbf{Z};\mathbf{\Theta},\mathbf{w}_a) = \frac{1}{1+\exp{\big(-a_{\mathbf{x}}(\mathbf{Z};\bm{\Theta},\mathbf{w}_a)\big)}},\label{eq:att_prob}
\end{equation}
then minimize the cross-entropy loss on the training data with per-voxel annotations:
\begin{align}
\ell_{\text{att}}(\mathbf{Z}^l,\mathbf{Y}^l;\mathbf{\Theta},\mathbf{w}_a)& = -\sum_{\mathbf{x}\in\mathcal{X}} \big[ y^l_\mathbf{x}\log{p^a_\mathbf{x}(\mathbf{Z}^l;\mathbf{\Theta},\mathbf{w}_a)}\nonumber \\
&+(1-y^l_\mathbf{x})\log(1-\big(p^a_\mathbf{x}(\mathbf{Z}^l;\mathbf{\Theta},\mathbf{w}_a)\big)\big], \label{eq:att_loss}
\end{align}
where $y^l_\mathbf{x}$ is the annotation of the instance at spatial location $\mathbf{x}$ obtained directly from $\mathbf{Y}^l$ and $\mathcal{X}$ is the whole $H'\times W'\times L'$ location lattice.

\subsection{Global Classifier}
\label{sec:attention_MIL}

We now introduce our attention-guided MIL module for image-level label prediction, \emph{i.e.}, normal/PDAC classification. MIL pooling usually plays an important role to form bag representations from {the} instance features. Typical MIL pooling algorithms include {{max}} pooling and {{average}} or {{sum}} pooling. But, {{max}} pooling can be easily influenced by noises, and {{average}} pooling or {{sum}} pooling may filter out PDAC signals. To address this problem, inspired by attention-based MIL \cite{Ilse18attention}, we use a weighted {average} pooling method, where the weights are determined by the attention values. First, we normalize the attention values (Eq.~\ref{eq:att_value}) over spatial locations to be attention weights by a softmax function:
\begin{equation}
    \alpha_\mathbf{x}(\mathbf{Z};\mathbf{\Theta},\mathbf{w}_a) = \frac{\exp\left(a_{\mathbf{x}}(\mathbf{Z};\bm{\Theta},\mathbf{w}_a)\right)}{\sum_{\mathbf{x}\in\mathcal{X}}\exp\left(a_{\mathbf{x}}(\mathbf{Z};\bm{\Theta},\mathbf{w}_a)\right)}.
\end{equation}
Note that, Eq.~\ref{eq:att_value} is generic for both $\mathbf{Z}^l$ and $\mathbf{Z}^u$. Then, we obtain the bag vector representation $\mathbf{h}(\mathbf{Z};\mathbf{\Theta}, \mathbf{w}_a)\in\mathbb{R}^C$ of image $\mathbf{Z}$ by our attention-guided MIL pooling:
\begin{equation}
\label{eq:bag_feature}
    \mathbf{h}(\mathbf{Z};\mathbf{\Theta}, \mathbf{w}_a) = \sum_{\mathbf{x}\in\mathcal{X}} \alpha_\mathbf{x}(\mathbf{Z};\mathbf{\Theta},\mathbf{w}_a)\mathbf{f}_\mathbf{x}(\mathbf{Z}; \mathbf{\Theta}).
\end{equation}
Based on the bag representation, we train a global image-level classifier for normal/PDAC classification by minimizing the cross-entropy loss:
\begin{align}
    \ell_\mathbbm{G}(\mathbf{Z},\mathbbm{y};\mathbf{\Theta},\mathbf{w}_a, \mathbf{w}_g) = -\big[\mathbbm{y}\log P_g(\mathbf{Z};\mathbf{\Theta},\mathbf{w}_a, \mathbf{w}_g)\nonumber\\
    + (1-\mathbbm{y})\log(1-P_g\big(\mathbf{Z};\mathbf{\Theta},\mathbf{w}_a, \mathbf{w}_g)\big)\big], \label{eq:mil_loss}
\end{align}
where $\mathbf{w}_g\in\mathbbm{R}^C$ is the parameter of the global image-level classifier, and
\begin{equation}
    P_g(\mathbf{Z};\mathbf{\Theta},\mathbf{w}_a, \mathbf{w}_g)=\frac{1}{1+\exp\big(-\mathbf{w}_g^{\top}\mathbf{h}(\mathbf{Z};\mathbf{\Theta},\mathbf{w}_a)\big)}, \label{eq:global_prob}
\end{equation}
representing the probability that image $\mathbf{Z}$ contains PDAC masses. Note that, the parameter $\mathbf{w}_a$ for computing attention values is optimized by minimizing both Eq.~\ref{eq:att_loss} and Eq.~\ref{eq:mil_loss}. So for both $\mathbf{Z}^l$ and $\mathbf{Z}^u$, their attention values are implicitly learned by minimizing Eq.~\ref{eq:mil_loss}, while the attention values of $\mathbf{Z}^l$ are explicitly learned by minimizing Eq.~\ref{eq:att_loss}, additionally. The loss function for training the global classifier over the whole trainig set $\mathcal{D}$ is defined by
\begin{align}\label{eq:loss_global}
L_{\text{global}}(\mathcal{D};\bm{\Theta},\mathbf{w}_a,\mathbf{w}_g)=\frac{1}{|\mathcal{D}|}\sum_{(\mathbf{Z},\mathbbm{y})\in\mathcal{D}}\ell_{{\mathbbm{G}}}(\mathbf{Z},\mathbbm{y};\mathbf{\Theta},\mathbf{w}_a, \mathbf{w}_g).
\end{align}

\subsection{Local Classifier}
\label{sec:local_classifier}
In this section, we describe how to use attention to guide semi-supervised PDAC segmentation. The key to improving segmentation results under the SSL setting is how to generate reliable pseudo-labels on the large amount of training data without per-voxel annotations. The attention values for $\mathbf{Z}^u$ computed by Eq.~\ref{eq:att_value} can be used as the per-voxel pseudo-labels after binarization. But, they are implicitly supervised by image-level annotation $\mathbbm{y}^u$. Consequently, they are obtained by searching over all spatial locations, and thus are coarse and noisy. Since many PDACs occupy only a small portion of pancreas regions, this issue becomes more severe. 

{Our basic strategy is to generate bag-level pseudo-labels to the data without per-voxel annotations, and then learn the local classifier based on these bag-level pseudo-labels by MIL.}
The bag-level pseudo-labels are obtained by separating the whole location lattice $\mathcal{X}$ into two bags (regions) according to the attention values: the PDAC bag $\mathcal{X}_f$ and the background bag $\mathcal{X}_b$.  This strategy leads to a benefit: The correctness of the bag-level pseudo-labels is much easier to be guaranteed than per-voxel pseudo-labels, \emph{e.g.}, a PDAC bag covers at least parts of PDAC masses and most of {the} instances in a background bag come from the background region (see Fig.~\ref{Fig:attention_example}).

\subsubsection{Attention based PDAC and Background Separation}
\label{subsec:separation}
$\mathcal{X}_f$ and $\mathcal{X}_b$ can be obtained by setting an attention threshold $t_a$:
$\mathcal{X}_f=\{\mathbf{x}\in\mathcal{X}|p_\mathbf{x}\leq t_a\}$ and $\mathcal{X}_b=\{\mathbf{x}\in\mathcal{X}|p_\mathbf{x}> t_a\}$. Here, $p_\mathbf{x}=p_\mathbf{x}(\mathbf{Z}^u;\mathbf{\Theta},\mathbf{w}_a)$ is computed by Eq.~\ref{eq:att_prob}. We omit these parameters for notational simplicity. Intuitively, the instances within each bag $\mathcal{X}_f$ or $\mathcal{X}_b$ should have similar attention values, while the instances from the two bags respectively should have different attention values. Thus, we determine the threshold $t_a$ for each volume $\mathbf{Z}^u$ adaptively by minimizing 
\begin{equation}
\label{eq:loss_clustering}
\ell_\mathbbm{C} = \sum_{c\in\{f,b\}}\sum_{\mathbf{x}\in\mathcal{X}_c}\|p_\mathbf{x}-\frac{1}{|\mathcal{X}_c|}\sum_{\mathbf{x}\in\mathcal{X}_c}p_\mathbf{x}\|.
\end{equation}
We realize this minimization by K-means clustering, since $\ell_\mathbbm{C}$ is equivalent to a K-means clustering loss when $K=2$. Note that, this minimization is performed in the forward pass, and does not participate in the backward pass.

\subsubsection{Learning Local Instance-level Classifier by MIL}
\label{subsec:local_mil}
We treat both the PDAC region and the background region as small bags: $\mathbf{Z}_f^u$ and $\mathbf{Z}_b^u$, and express the bag vector representation $\bm{\hbar}(\mathbf{Z}_c^u;\mathbf{\Theta})\in\mathbb{R}^C$ by the aggregation of the instance features:
\begin{equation}
    \bm{\hbar}(\mathbf{Z}_c^u;\mathbf{\Theta},\mathbf{w}_a)=\frac{1}{|\mathcal{X}_c|}\sum_{\mathbf{x}\in\mathcal{X}_c} \mathbf{f}_\mathbf{x}(\mathbf{Z}^u;\mathbf{\Theta},\mathbf{w}_a), c\in\{f,b\}.\label{eq:bag_fore_back}
\end{equation}
Note that, $\mathcal{X}_c$ depends on parameter $\mathbf{w}_a$. Hence, the left hand side (LHS) of Eq.~\ref{eq:bag_fore_back} also has parameter $\mathbf{w}_a$.  
Here we use {average} pooling, because we want the PDAC bags and background bags to be compact, and to progressively approach ground-truth segmentation maps. 
 
The probability that a bag is a PDAC region is defined by:
\begin{equation}
P_c(\mathbf{Z}^u_c;\mathbf{\Theta},\mathbf{w}_a,\mathbf{w}_l) = \frac{1}{1+\exp\big(-\mathbf{w}_l^{\top}\bm{\hbar}(\mathbf{Z}^u_c;\mathbf{\Theta},\mathbf{w}_a)\big)}, \label{eq:bag_prob}   
\end{equation}
where $\mathbf{w}_l\in\mathbbm{R}^C$ is the parameter of the local classifier. Then an MIL loss is defined on $\mathbf{Z}^u$ to learn the local classifier:
\begin{align}
\ell^u_\mathbbm{L}(\mathbf{Z}^u;\bm{\Theta},\mathbf{w}_a,\mathbf{w}_l)=-\big[\mathbbm{y}^u_f\log{P_c(\mathbf{Z}^u_f;\mathbf{\Theta},\mathbf{w}_a,\mathbf{w}_l)}\nonumber \\+(1-\mathbbm{y}^u_b)\log\big(1-P_c(\mathbf{Z}^u_b;\mathbf{\Theta},\mathbf{w}_a,\mathbf{w}_l)\big)\big].\label{eq:SSL_unlabeled}
\end{align}
Note that, since $\mathbf{w}_l$ are linear coefficients, we have:
\begin{equation}
    \mathbf{w}_l^{\top}\bm{\hbar}(\mathbf{Z}^u_c;\mathbf{\Theta},\mathbf{w}_a)=\frac{1}{|\mathcal{X}_c|}\sum_{\mathbf{x}\in\mathcal{X}_c} \mathbf{w}_l^{\top}\mathbf{f}_\mathbf{x}(\mathbf{Z}^u;\mathbf{\Theta},\mathbf{w}_a),
\end{equation}
{Since the bag-level feature is the average of the instance-level features and the bag-level classifier is linear, applying the linear coefficients to the bag-level feature (the bag score) is equivalent to applying the linear coefficients to each instance-level feature (the instance score) then averaging. This implies that the local classifier can be directly applied to an instance.} With $\mathbf{w}_l$, we have a unified function to compute the probability that an instance $\mathbf{f}_\mathbf{x}(\mathbf{Z};\mathbf{\Theta})$, no matter from $\mathbf{Z}^l$ or $\mathbf{Z}^u$, belongs to a PDAC region:
\begin{equation}
    q_\mathbf{x}(\mathbf{Z};\mathbf{\Theta},\mathbf{w}_l) = \frac{1}{1+\exp\big(-\mathbf{w}_l^{\top}\mathbf{f}_\mathbf{x}(\mathbf{Z};\mathbf{\Theta})\big)}.\label{eq:local_prob}
\end{equation}
{Eq.~\ref{eq:bag_prob} and Eq.~\ref{eq:local_prob} show a unified function to compute the probabilities for both a bag and an instance by using the local classifier $\mathbf{w}_l$.} Then, we can also define a loss function on the training data with per-voxel annotations to learn the local classifier:
\begin{align}
    \ell^l_{\mathbbm{L}}(\mathbf{Z}^l,\mathbf{Y}^l;\mathbf{\Theta},\mathbf{w}_l) &= -\sum_{\mathbf{x}\in\mathcal{X}}\big[y_{\mathbf{x}}^l\log q_\mathbf{x}(\mathbf{Z}^l;\mathbf{\Theta},\mathbf{w}_l)\nonumber\\
    &+(1-y_{\mathbf{x}}^l)\log \big(1-q_\mathbf{x}(\mathbf{Z}_l;\mathbf{\Theta},\mathbf{w}_l)\big)\big].\label{eq:SSL_labeled}
\end{align}

Now, by combing Eq.~\ref{eq:SSL_unlabeled} and Eq.~\ref{eq:SSL_labeled}, we can write down the loss function over the training set $\mathcal{D}$ to learn the local instance-level classifier:
\begin{align}
L_{\mathbbm{L}}(\mathcal{D};\bm{\Theta},\mathbf{w}_a,&\mathbf{w}_l)=\frac{1}{N_p^l}\sum_{(\mathbf{Z}^l,\mathbf{Y}^l)\in\mathcal{D}^l}\mathbbm{y}^l\ell^l_\mathbbm{L}(\mathbf{Z}^l,\mathbf{Y}^l;\mathbf{\Theta},\mathbf{w}_l)+\nonumber \\&\frac{\lambda}{N_p^u}\sum_{(\mathbf{Z}^u,\mathbbm{y}^u)\in\mathcal{D}^u}\mathbbm{y}^u\ell^u_\mathbbm{L}(\mathbf{Z}^u;\bm{\Theta},\mathbf{w}_a,\mathbf{w}_l),\label{eq:SSL}
\end{align}
where $N_p^l=\sum_{(\mathbf{Z}^l,\mathbf{Y}^l)\in\mathcal{D}^l}\mathbbm{y}^l$ and $N_p^u=\sum_{(\mathbf{Z}^u,\mathbbm{y}^u)\in\mathcal{D}^u}\mathbbm{y}^u$ are the numbers of the training data which have PDAC masses, with and without per-voxel annotations, respectively. Since the local classifier is used for PDAC segmentation, we learn it on the training data which have PDAC masses. The training data which have no PDAC mass are excluded for learning local classifier, as shown in Eq.~\ref{eq:SSL}. $\lambda$ is a weight factor to balance the two terms of the right hand side (RHS) of Eq.~\ref{eq:SSL}. Intuitively, at the beginning of training, the pseudo PDAC region $\mathcal{X}_f$ and the background region $\mathcal{X}_b$ are very noisy, which makes the second term of the RHS of Eq.~\ref{eq:SSL} unreliable. As the optimization proceeds, it progressively become reliable. Here, we make use of the attention guidance again, to design an adaptive weight factor to automatically reflect the reliability of the second term: $\lambda = \max_{\mathbf{x}\in\mathcal{X}} p_\mathbf{x}(\mathbf{Z}_u;\mathbf{\Theta},\mathbf{w}_a)$.

Since the explicitly-learned attention is a proxy of the local classifier, we add its loss function over the training set $\mathcal{D}$ to Eq.~\ref{eq:SSL}, which obtains the final loss function to learn the local classifier:
\begin{equation}
  L_{\text{local}}(\mathcal{D};\bm{\Theta},\mathbf{w}_a,\mathbf{w}_l)=L_{\text{att}}(\mathcal{D}^l;\bm{\Theta},\mathbf{w}_a) + L_{\mathbbm{L}}(\mathcal{D};\bm{\Theta},\mathbf{w}_a,\mathbf{w}_l),
\end{equation}
where 
 \begin{align}
L_{\text{att}}(\mathcal{D}^l;\bm{\Theta},\mathbf{w}_a)=\frac{1}{N_p^l}\sum_{(\mathbf{Z}^l,\mathbf{Y}^l)\in\mathcal{D}^l}\mathbbm{y}^l\ell_{\text{att}}(\mathbf{Z}^l,\mathbf{Y}^l;\mathbf{\Theta},\mathbf{w}_a).
\vspace{-1em}
\end{align}

\subsubsection{Overall Loss Function}
Finally, we write down the overall loss function for training our IAG-Net:
\begin{align}
L_{\text{IAG}}(&\mathcal{D};\bm{\Theta},\mathbf{w}_a,\mathbf{w}_g,\mathbf{w}_l)=\nonumber\\&L_{\text{global}}(\mathcal{D};\bm{\Theta},\mathbf{w}_a,\mathbf{w}_g)+\beta L_{\text{local}}(\mathcal{D};\bm{\Theta},\mathbf{w}_a,\mathbf{w}_l),\label{eq:overall_loss}
\end{align}
where $\beta$ is a trade-off parameter which balances
the two terms (we set $\beta$ = 20 in our implementation, which is not sensitive between [10, 30]). All parameters are jointly optimized during network training. The optimized parameters are obtained by
\begin{align}
    &(\bm{\Theta},\mathbf{w}_a,\mathbf{w}_g,\mathbf{w}_l)^\ast=\nonumber\\
    &\arg\min_{\bm{\Theta},\mathbf{w}_a,\mathbf{w}_g,\mathbf{w}_l}L_{\text{IAG}}(\mathcal{D};\bm{\Theta},\mathbf{w}_a,\mathbf{w}_g,\mathbf{w}_l).
\end{align}

\begin{figure*}[t]
\begin{center}
    \includegraphics[width=0.8\linewidth]{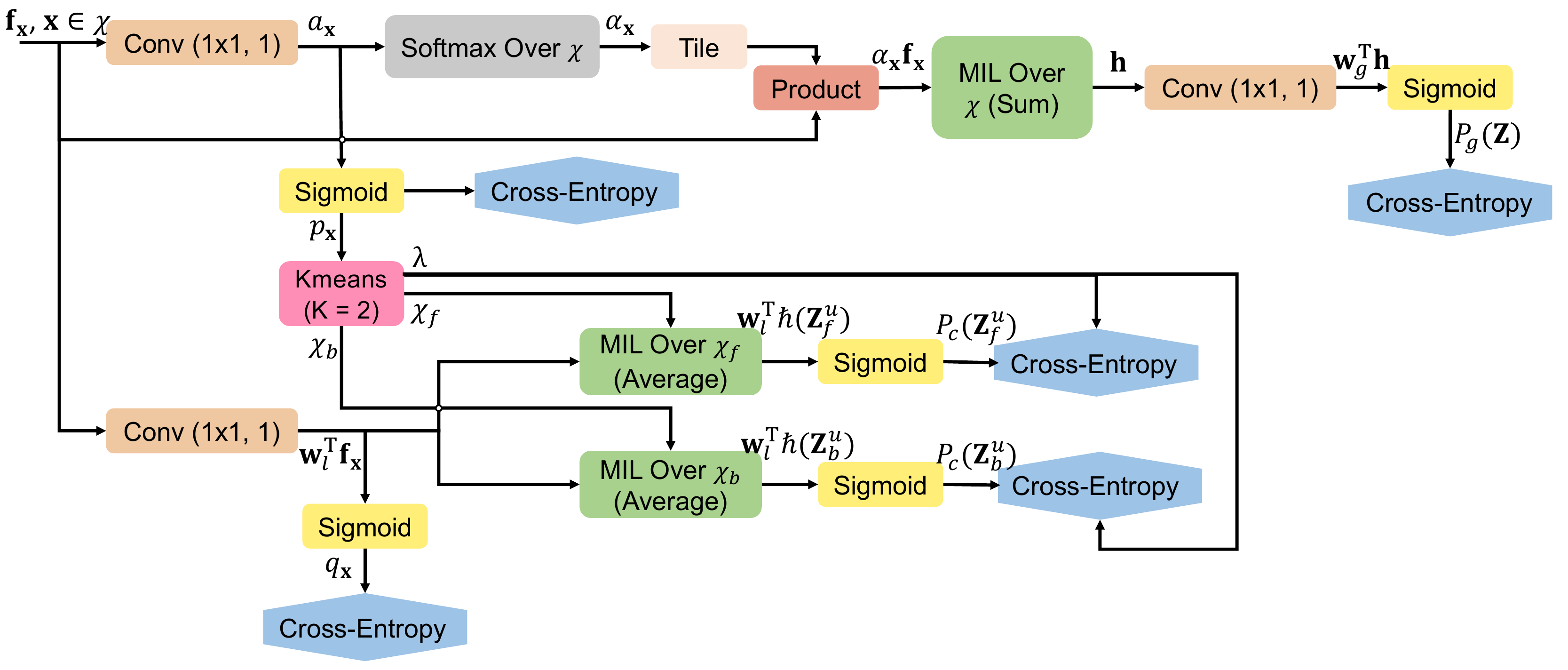}
\end{center}
\caption{{The detailed architecture shown layer by layer of our IAG-Net. The backbone is not illustrated for simplicity.}
} 
\label{Fig:architecture}
\end{figure*}

The overall training procedure is summarized in Algorithm~\ref{Alg:train}. {We also show the detailed architecture of our IAG-Net in Fig. \ref{Fig:architecture}. The backbone is not illustrated for simplicity.}

\begin{algorithm}[t!]
\small
\SetKwInOut{Input}{Input}
\SetKwInOut{Output}{Output}
\SetKwInOut{Return}{Return}
\Input{
Training set $\mathcal{D}=\mathcal{D}^l \bigcup \mathcal{D}^u$, where $\mathcal{D}^l=$ $\{(\mathbf{Z}^l,\mathbf{Y}^l)\}_{l=1}^L$ and $\mathcal{D}^u = \{(\mathbf{Z}^u,\mathbbm{y}^u)\}_{u=L+1}^U$;\\
Max number of iterations $T$;\\
}
\Output{
Parameters $\bm{\Theta}^*$, $\mathbf{w}_a^*$, $\mathbf{w}_l^*$ and $\mathbf{w}_g^*$;
}
${t}\leftarrow{0}$;\\
Randomly initialize $\bm{\Theta}$, $\mathbf{w}_a$, $\mathbf{w}_l$ and $\mathbf{w}_g$;\\
\Repeat{${t}={T}$}
{${t}\leftarrow{t+1}$;\\
Randomly select a data sample $(\mathbf{Z},\cdot)$ from $\mathcal{D}$\\
\eIf{$(\mathbf{Z},\cdot)\in\mathcal{D}^l$}
{Compute $\ell_{\text{att}}(\mathbf{Z}^l,\mathbf{Y}^l;\mathbf{\Theta},\mathbf{w}_a)$ by Eq.~\ref{eq:att_loss} and $\ell^l_{\mathbbm{L}}(\mathbf{Z}^l,\mathbf{Y}^l;\mathbf{\Theta},\mathbf{w}_l)$ by Eq.~\ref{eq:SSL_labeled};\\ }
{Obtain $\mathcal{X}_f$ and $\mathcal{X}_b$ by minimizing Eq.~\ref{eq:loss_clustering}\\
Compute $\ell^u_\mathbbm{L}(\mathbf{Z}^u;\bm{\Theta},\mathbf{w}_a,\mathbf{w}_l)$ on $\{\mathcal{X}_f, \mathcal{X}_b\}$ by Eq.~\ref{eq:SSL_unlabeled};\\ }
Compute $\ell_{\mathbbm{G}}(\mathbf{Z},\mathbbm{y};\mathbf{\Theta},\mathbf{w}_a, \mathbf{w}_g)$ by Eq.~\ref{eq:mil_loss};\\ 
Compute $L_{\text{IAG}}(\mathcal{D};\bm{\Theta},\mathbf{w}_a,\mathbf{w}_g,\mathbf{w}_l)$ by Eq.~\ref{eq:overall_loss}; \\
Update $\bm\Theta$, $\mathbf{w}_a$, $\mathbf{w}_l$ and $\mathbf{w}_g$ by Gradient Descent;
}
\Return{
$(\bm\Theta, \mathbf{w}_a, \mathbf{w}_l, \mathbf{w}_g)^*\leftarrow{(\bm\Theta, \mathbf{w}_a, \mathbf{w}_l, \mathbf{w}_g)}$.
}
\caption{
The training process of IAG-Net
}
\label{Alg:train}
\end{algorithm}
\setlength{\textfloatsep}{6pt}

\subsection{Testing IAG-Net for PDAC Prediction}
\label{sec:testing}
Given a testing image $\mathbf{Z}$, whether it has PDAC masses is determined by the global image-level classifier (Eq.~\ref{eq:global_prob}):
\begin{equation}
\label{eq:image_pred}
    \hat{\mathbbm{y}}=\left\{\begin{matrix}
1, & \text{if } P_g(\mathbf{Z};\mathbf{\Theta}^\ast,\mathbf{w}_a^\ast, \mathbf{w}_g^\ast)\geqslant 0.5;\\ 
0, & \text{otherwise}.
\end{matrix}\right.
\end{equation}
If $ \hat{\mathbbm{y}}=1$, then we further need to segment the PDAC masses out from $\mathbf{Z}$, \emph{i.e.}, to predict $\hat{y}_{\mathbf{x}}$ at each location $\mathbf{x}$. For this task, we have the result of the local instance-level classifier (Eq.~\ref{eq:local_prob}): $q_\mathbf{x}(\mathbf{Z};\mathbf{\Theta}^\ast,\mathbf{w}_l^\ast)$. Recall that, our learned attention guidance is inductive. It acts as a proxy of a local instance-level classifier (Eq.~\ref{eq:att_prob}): $p_\mathbf{x}(\mathbf{Z};\mathbf{\Theta}^\ast,\mathbf{w}_a^\ast)$. We simply average these two results to determine the value of $\hat{y}_{\mathbf{x}}$:
\begin{equation}
\label{eq:voxel_pred}
    \hat{y}_\mathbf{x}=\left\{\begin{matrix}
1, & \text{if}~ p_\mathbf{x}(\mathbf{Z};\mathbf{\Theta}^\ast,\mathbf{w}_a^\ast)+q_\mathbf{x}(\mathbf{Z};\mathbf{\Theta}^\ast,\mathbf{w}_l^\ast) \geqslant 1.0;\\ 
0, & \text{otherwise}.
\end{matrix}\right.
\end{equation}

\section{Experimental Results}
In this section we describe the implementation details and
compare IAG-Net with other competitors. We first evaluate our approach in a JHMI dataset which consists of both normal and PDAC cases, and then provide some diagnostic experiments for further analysis. After that, we apply our approach to segment pancreatic tumors in a public MSD challenge dataset under the semi-supervised setting.

\subsection{Dataset}
We evaluate IAG-Net in a JHMI dataset, which contains $400$ CT images of normal cases and $400$ biopsy-proven PDAC cases under an IRB (Institutional Review Board) approved protocol in Johns Hopkins Hospital as a part of the FELIX project for pancreatic
cancer research. CT images have voxel spatial resolution of $([0.523 - 0.977] \times [0.523 - 0.977] \times 0.5) mm^3$. All CT scans are pancreas region of interest (RoI) of contrast enhanced images in {the} Venous phase. 
Each PDAC case has tumor annotation in biopsy-proven PDAC cases, annotated and verified by an experienced board-certified Abdominal Radiologist. We randomly partition the dataset into 4 equally-sized folds, using three of them for training, and the remaining one for testing. Unless otherwise specified, half of the biopsy-proven PDAC cases are randomly chosen to be given only voxel-level annotations during training. We also evaluate the generality of IAG-Net in a public MSD pancreas tumor dataset \cite{Simpson19a}. This dataset was comprised of patients undergoing resection of pancreatic masses (intraductal mucinous neoplasms, pancreatic neuroendocrine tumours, or PDAC).

\subsection{Evaluation metric}
For PDAC segmentation, Dice-S{\o}rensen similarity coefficient (DSC) over the whole pancreas RoI is computed, and mean DSC scores, standard deviation, max DSC and median DSC over all testing cases are reported. For normal/PDAC classification, sensitivity and specificity are calculated.

\subsection{Implementation details}
{For the data pre-processing, we truncate the raw intensity values within the range [-125, 350] HU and normalize the intensity values of each raw CT case to [0, 1] HU to decrease the data variance caused by the physical processes of medical images \cite{Zhu2018a}.} In the experiment, we test our IAG-Net based on different popular CNN backbones: VGG-Net \cite{simonyan2015very} and 2D U-Net \cite{Ronneberger2015u}. Given an input 3D volume $\mathbf{Z}$, we use the CNN backbone to compute a feature map of each slice, and then concatenate all the feature maps as the feature map $\mathbf{F}$ of the volume $\mathbf{Z}$.
{We choose the output of the last feature extraction layer in both backbones as the feature map $\mathbf{F}$. For VGG-Net, extracting feature maps from \texttt{conv6} will cause \emph{out-of-memory} errors {on an NVidia TITAN RTX GPU with 24GB of GDDR6 memory}. So we use the output of \texttt{conv5} as our feature map $\mathbf{F}$, and $C=512$}. The per-voxel annotation is downsampled to match the dimension of $\mathbf{F}$. The predicted per-voxel label map is upsampled to the original size. For 2D U-Net, we use the output of the layer before the last as our feature map $\mathbf{F}$, and $C=64$. The dimension of $\mathbf{F}$ matches with the input image.

Noted that feeding 2D slices from different viewing directions into 2D deep networks, and combining segmentation results from multi-view images as the final prediction leads to comparable performance to 3D deep networks in medical image segmentation \cite{Wang2019abdominal,Xie2019Recurrent}. In addition, radiologists also view 3D scans slice-by-slice when they do annotation. So we simply adopt 2D deep networks rather than 3D networks. Our IAG-Net can be also built on 3D backbones, but this is out of the scope of this paper.

Given a 3D pancreas RoI, we train IAG-Nets on 2D views based on the normal vector directions of the sagittal (X), coronal (Y) and axial (Z) planes, respectively. When training an IAG-Net for one direction, {\em{e.g.}}, Z plane, in order to save memory, instead of feeding all slices in the whole 3D volume into the IAG-Net as one batch, we sample slices with an interval which is a random integer between $[1,5]$, so that the size of the real input is $\hat{H}\times W \times L$ where $\hat{H}<H$. The same strategy is also applied to other directions. Following~\cite{Zhou2019semi,Xie2019Recurrent}, the final prediction, including classification and segmentation, is a combination of three 2D views. {These strategies are applied for other competing methods, unless otherwise specified.}

{VGG-Net was pre-trained on ImageNet, and it was trained on the PascalVOC dataset to transfer the learned weights from classification networks to segmentation networks.} U-Net was pre-trained on a separate in-house multi-organ segmentation dataset. These pre-trained models are used for the rest of the experiments. We set the initial learning rate to be $10^{-5}$ for VGG-Net and $10^{-4}$ for U-Net. {Models are trained for $120,000$ iterations. We use exponential learning rate decay with $\gamma = 0.99$.} {The learning rate, maximum training iterations, and learning rate decay for training are the same for other competing methods unless otherwise specified. }

\begin{table*}[t!]
\small
\definecolor{Gray}{gray}{0.9}
\centering
\caption{Performance comparison ($\%$) on PDAC segmentation (DSC, mean $\pm$ standard deviation, max and median of all cases) and classification (sensitivity and specificity). {As a reference, we also show performances of using all the training data in a fully-supervised manner as IAG-Net (fully-supervised).} \textbf{Bold} denotes the best results {of semi-supervised methods}. {Number of parameters of the networks, significant statistical improvements of IAG-Net vs. FCN-8s \cite{long2015fully}/U-Net \cite{Ronneberger2015u}, Li \emph{et al.} \cite{Li2018thoracic}, DMPCT \cite{Zhou2019semi}, Collaborative learning method \cite{Zhou2019collaborative}{, and statistical improvements of IAG-Net (fully-supervised) vs. IAG-Net} on PDAC segmentation are shown.}}
\label{Tab:CompareBase}
\begin{tabular}{llccccc}
\toprule[0.15em]
Backbone & {Method}  & Mean DSC (Max, Median) & Sensitivity & Specificity & { \# Parameters} & $p$-value\\
\midrule[0.1em]
\multirow{7}{*}{VGG-Net} & Attention-based MIL \cite{Ilse18attention} & $-$ & 99.00 & \textbf{98.00} & {14.78M} & {$-$} \\
& {FCN-8s \cite{long2015fully}} & {50.02~$\pm$~26.15 (89.75, 56.50)} & {$-$} & {$-$} & {134.27M} & {{7.93$\times$10$^{-5}$}}\\
& Li {\em{et al.}} \cite{Li2018thoracic} & 41.31~$\pm$~21.41 (82.80, 41.55) & 99.00 & 96.75 & {14.72M} & {3.68$\times$10$^{-51}$}\\
& DMPCT \cite{Zhou2019semi}  & 49.24~$\pm$~27.04 (90.89, 54.12) & $-$ & $-$ & {134.27M} & {1.87$\times$10$^{-7}$}\\
& {Collaborative \cite{Zhou2019collaborative}} & {52.52~$\pm$~19.35 (85.88, 55.37)} & {98.25} & {96.75} & {21.02M} &{1.31$\times$10$^{-7}$} \\
& IAG-Net (Ours)  & {54.38}~$\pm$~{18.77} (87.65, 57.00) & 99.25 & 97.50 & {14.72M} & {$-$}\\
\cmidrule(lr){2-7}
& {IAG-Net (fully-supervised)} & {55.45~$\pm$~19.60 (88.36, 58.29)} & {99.25} & {97.75} & {14.72M} & {3.90$\times$10$^{-3}$}\\
\midrule
\multirow{7}{*}{U-Net} & Attention-based MIL \cite{Ilse18attention} & $-$ & 97.00 & 94.50 & {30.44M} & {$-$} \\
& {U-Net \cite{Ronneberger2015u}} & {51.87~$\pm$~25.94 (93.63, 56.52)} & {$-$} & {$-$} & {30.43M} & {{1.22$\times$10$^{-32}$}}\\
& Li {\em{et al.}} \cite{Li2018thoracic} & 47.91~$\pm$~26.13 (90.84, 51.73) & 99.25 & 93.75 & {30.43M} & {7.70$\times$10$^{-45}$}\\
& DMPCT \cite{Zhou2019semi} & 52.35~$\pm$~26.38 (92.23, 56.69) & $-$ & $-$ & {30.43M} & {9.61$\times$10$^{-30}$}\\
& {Collaborative \cite{Zhou2019collaborative}} & {55.24~$\pm$~24.96 (93.88, 60.94)} & {98.75} & {95.75} & {36.74M} & {3.70$\times$10$^{-18}$}\\
& IAG-Net (Ours) & \textbf{60.29}~$\pm$~21.60 (\textbf{94.04}, \textbf{64.37}) & \textbf{99.75} & 96.50 & {30.43M} & {$-$} \\
\cmidrule(lr){2-7}
& {IAG-Net (fully-supervised)} & {60.38~$\pm$~23.83 (93.61, 66.93)} & {98.25} & {98.00} & {30.43M} & {6.16$\times$10$^{-1}$}\\
\bottomrule[0.15em]
\end{tabular}
\vspace{-0.8em}
\end{table*}

\subsection{Comparison between IAG-Net and {Other Methods}}

In this section, we conduct comparison between IAG-Net and {four competitors}: 1) attention-based MIL \cite{Ilse18attention} 2) Li {\em{et al.}} \cite{Li2018thoracic}, which is a unified framework to combine disease identification and localization of abnormalities, 3) a semi-supervised learning method for segmentation, {\em{i.e.}}, Deep Multi-Planar Co-Training (DMPCT) \cite{Zhou2019semi}, {and 4) a collaborative learning method for segmentation and classification under the semi-supervised setting \cite{Zhou2019collaborative}. Besides, we also compare with the segmentation backbones, \emph{i.e.}, directly using FCN-8s \cite{long2015fully} and U-Net for segmentation. Noted that the backbone of FCN-8s is VGG-Net, but it adopts additional strategies such as skip connections to enhance the segmentation performance.}

Attention-based MIL \cite{Ilse18attention} uses only image-level annotation during training. But, it shows the ability to do segmentation {\em{i.e.}}, there is a substantial matching between the heat map obtained from attention and the segmentation ground-truth. We use attention-based MIL as a baseline method. Following \cite{Ilse18attention}, we treat $a_\mathbf{x}' = (a_\mathbf{x} - \min(\mathbf{a}))/(\max(\mathbf{a}) - \min(\mathbf{a}))$ as the instance probability to obtain the PDAC segmentation results.

The work of Li {\em{et al.}} \cite{Li2018thoracic} is designed for disease identification and localization for 2D images, given part of the images with bounding boxes annotations and all images have disease identities. Since our task is to perform segmentation on 3D volumes, we customize Li's method to fit our setting: We replace the MIL {xor} operation used in Li's method by the commonly used MIL {max} pooling, since the number of instances of a 3D volume {is} 50$\times$ more than that of a 2D image, which makes MIL based on {xor} operation difficult to converge, even with the smoothing trick ({\em{i.e.}}, used in Li's method, normalize the patch score from $[0, 1]$ to $[0.98, 1]$). We also change the loss function in Li's method to the same cross-entropy loss as ours for PDAC segmentation and replace the backbone utilized in Li's method with our backbones.

In implementing DMPCT \cite{Zhou2019semi}, We also adopt VGG-Net and U-Net as backbones. When using VGG-Net as its backbone, followed by \cite{Zhou2019semi}, we also use the strategies in FCN-8s~\cite{long2015fully}, such as skip connections and fusing coarse and fine feature maps \cite{long2015fully} to improve its segmentation results \cite{Zhou2019semi}. {We follow the same settings as reported in \cite{Zhou2019semi}, \emph{i.e.}, we set the learning rate to be $10^{-9}$ for FCN-8s. The teacher model and the student model are trained for 80,000 and 160,000 iterations, respectively for both FCN-8s and U-Net. For U-Net, we set the learning rate to be $10^{-8}$. Followed by \cite{Zhou2019semi}, models are trained and tested from multiple planes separately in a slice-by-slice manner. The final prediction is a combination of three 2D views.}

{To implement the collaborative learning method \cite{Zhou2019collaborative}, we replace the lesion segmentation network {with} our backbones, and adopt the other network components (\emph{e.g.}, lesion attention classification) as illustrated in \cite{Zhou2019collaborative}. Since the number of our class is 2, we slightly modify the network accordingly.}

Results are shown in Table~\ref{Tab:CompareBase}. Attention-based MIL \cite{Ilse18attention} achieves high classification accuracy with VGG-Net, but does not work for PDAC segmentation at all. More specifically, attention values are uniformly distributed over all locations. It is no surprise to observe such a phenomenon, since without any per-voxel supervision, it is hard to learn the attention for our dataset. Li {\em{et al.}} \cite{Li2018thoracic} achieves comparable classification results with our IAG-Net using VGG-Net as the backbone, but it performs worse in PDAC segmentation compared to ours. We achieve much better segmentation and classification results than Li's method when using U-Net as the backbone model.
DMPCT \cite{Zhou2019semi} is designed only for segmentation. We can outperform DMPCT by 5.14\% and 7.94\% in terms of mean DSC with two backbones, respectively. Note that DMPCT mines consensus information from multiple planes, which is demonstrated to generate more reliable pseudo-labels than single-planar based method \cite{Bai2017semi}. Since our current method learns attention-guided pseudo-labels for each plane separately, more performance gain is expected if consensus information from multiple planes can be extracted. Moreover, DMPCT with VGG-Net adopts additional strategies from FCN-8s~\cite{long2015fully} to enhance the segmentation results, Our results can be further improved by using these strategies. {IAG-Net achieves superior performances than the collaborative learning method \cite{Zhou2019collaborative} in terms of both segmentation and classification tasks.} {We show the performance of a fully supervised IAG-Net in Table~\ref{Tab:CompareBase}, termed as IAG-Net (fully-supervised) for reference. The fully supervised IAG-Net is trained under the setting that all the training data are with per-voxel annotations, \emph{i.e.}, $|\mathcal{D}^l|=300$, $|\mathcal{D}^u|=0$, and Eq~(12) is omitted during training. The results show that the proposed IAG-Net can approach its fully-supervised version.} 

\begin{figure}[t]
\begin{center}
    \includegraphics[width=0.9\linewidth]{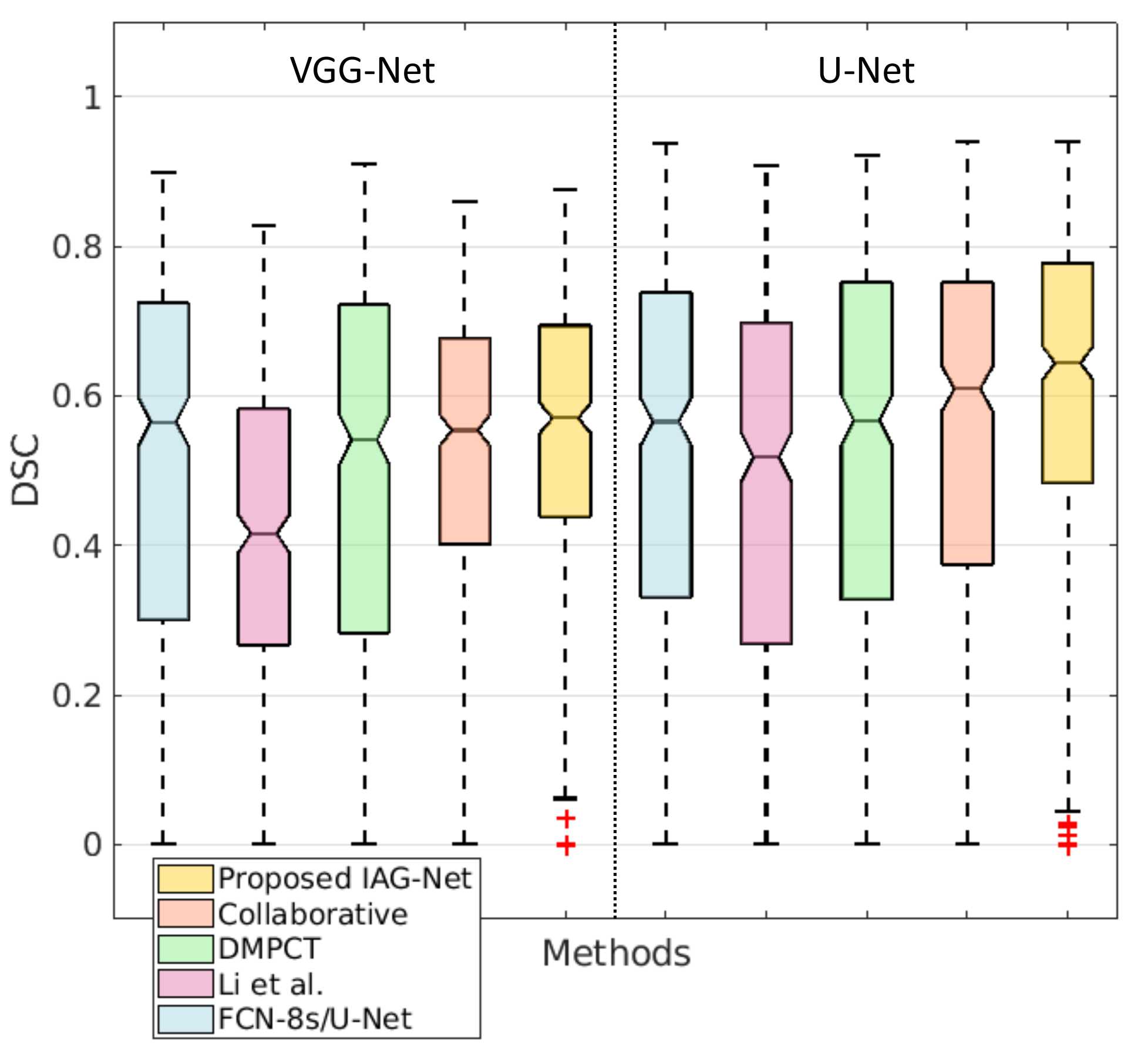}
\end{center}
\vspace{-0.5em}
\caption{Performance comparison (DSC) in box plots of PDAC segmentation. Proposed IAG-Net improves the overall mean and median DSC and also reduces the standard deviation.} 
\label{Fig:Boxplot}
\end{figure}

\begin{figure*}[t]
\begin{center}
    \includegraphics[width=0.9\linewidth]{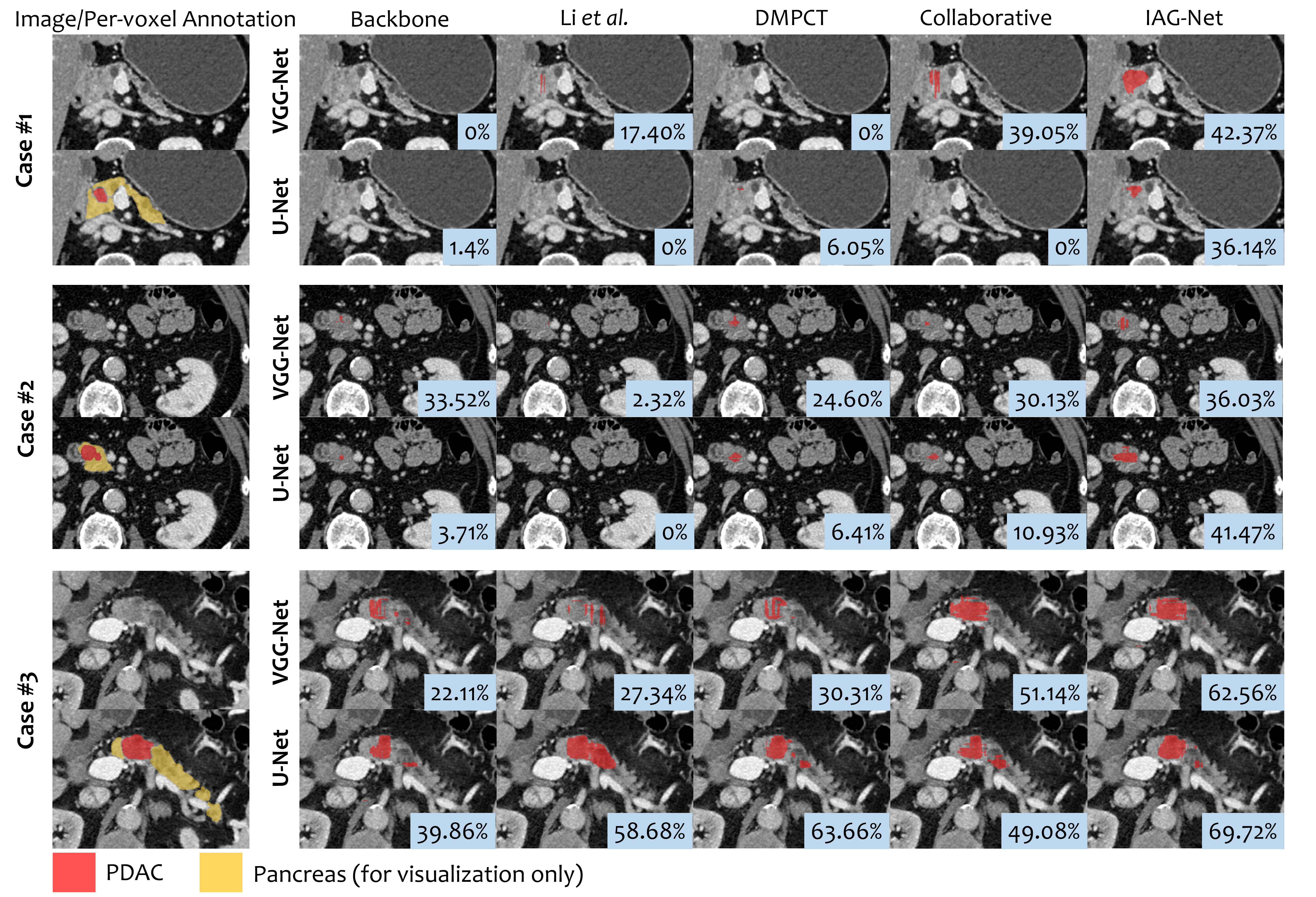}
\end{center}
\vspace{-0.7em}
\caption{Qualitative comparisons of PDAC segmentation. Three cases are shown in 2D slices in axial view. {For each case, we show the results obtained by} different methods ({\em{i.e.}}, {a backbone network (VGG-Net \cite{simonyan2015very} based FCN-8s \cite{long2015fully}, or U-Net \cite{Ronneberger2015u}),} Li {\em{et al.}} \cite{Li2018thoracic}, DMPCT \cite{Zhou2019semi}{, collaborative learning method \cite{Zhou2019collaborative} and proposed IAG-Net) with the two backbone models, \emph{i.e.}, VGG-Net (upper) and U-Net (lower).} Per-voxel pancreas annotation is also shown as reference. Numbers on the bottom right are segmentation DSCs.} 
\label{Fig:Vis}
\vspace{-1em}
\end{figure*}

Fig.~\ref{Fig:Boxplot} shows comparison results of our IAG-Net, {two backbone networks, Li \emph{et al.} \cite{Li2018thoracic}, DMPCT \cite{Zhou2019semi} and collaborative} {learning \cite{Zhou2019collaborative}} by box plots. Besides, the $p$-values for testing significant difference between {our IAG-Net and backbones, Li {\em{et al.}} , DMPCT and collaborative learning method} for PDAC segmentation are shown in the last column of Table~\ref{Tab:CompareBase}. Our IAG-Net is significantly better than {all} competitors in PDAC segmentation. { We compare parameter sizes among different methods. As shown in Table~\ref{Tab:CompareBase}, compared with other methods, our IAG-Net is effectively designed for PDAC prediction with a negligible increase in parameter sizes.} We also illustrate PDAC segmentation results in Fig.~\ref{Fig:Vis} for qualitative comparison. We can see that compared with {other methods}, IAG-Net can output more accurate segmentation results, which are more robust to the complicated background.

We compare with {two} state-of-the-art consistency-{related} {methods} whose codes are publicly available \cite{Ref:LiYCFH20,Ouali2020semi}, {achieving 57.22~$\pm$~21.68 (89.53, 64.22) and 51.21~$\pm$~24.11 (91.86, 55.24)} in terms of DSC for PDAC segmentation{, respectively}. {To run these methods, we use the same training models (backbone: DenseUNet \cite{Li2018H} for \cite{Ref:LiYCFH20} and ResNet-50 \cite{He2016deep} for \cite{Ouali2020semi}) and parameters as reported in \cite{Ref:LiYCFH20} and \cite{Ouali2020semi}. Models are trained and tested from multiple planes separately in a slice-by-slice manner. The final prediction is a combination of three 2D views.} 

{Note that, \cite{Ouali2020semi}} uses ResNet-50 as the backbone network, which is a stronger backbone than VGG-Net, {as shown in} \cite{He2016deep}. {We also test our IAG-Net with ResNet-50, and achieve 56.36~$\pm$~18.37 (91.90, 58.77). This shows that IAG-Net outperforms \cite{Ouali2020semi} by a large margin (with 5.15\% improvement in terms of mean DSC). Moreover, we adopt the cross-consistency training strategy in our IAG-Net (\cite{Ref:LiYCFH20} takes advantages of both transformation consistency and self-ensembling, while  \cite{Ouali2020semi} designs a {stand-alone}  cross-consistency training strategy which can be directly integrated into our framework), \emph{i.e.}, we add perturbations on the encoder's output, and add the auxiliary global classifier and the local classifier corresponding to each perturbation. More specifically, we use $K = 2$ for Con-Msk and Obj-Msk, $K = 2$ for I-VAT, and $K = 6$ for the rest of the perturbations, as suggested in \cite{Ouali2020semi}. With cross-consistency training, our IAG-Net-consistency can obtain 57.44~$\pm$~23.23 (92.92, 63.33), whose performance is better than pure IAG-Net or consistency-based method \cite{Ouali2020semi}. This shows} that the consistency-based methods are another direction for semi-supervised learning which are also demonstrated to be complementary with self-training methods in \cite{Tang2020proposal}.

Last but not the least, we briefly show the results of applying 3D U-Net \cite{Cicek163d} for only PDAC segmentation as a reference. {We set the initial learning rate to be $10^{-2}$, and use exponential learning rate decay with $\gamma=0.99$. During training, we randomly sample patches of a specified size (\emph{i.e.}, 64). During testing, we employ the sliding window strategy to obtain the final predictions.} The mean DSC (\%, max, median) is 50.13~$\pm$~26.75 (93.06, 56.64) with half of the PDAC cases with per-voxel annotation during training. Our IAG-Net performs much better than the 3D U-Net baseline for PDAC segmentation.

\subsection{Ablation Study}
We conduct ablation experiments to analyze the influence of different designs and components for IAG-Net. 

\noindent\textbf{(1) Ablation study on the joint learning framework.}

We first conduct an ablation study on the joint framework, \emph{i.e.}, to explore what if only the global or the local classifier is learned in IAG-Net. To train the global classifier only, we disable the supervision for the local classifier in IAG-Net; To train the local classifiers only, we disable the supervision for the global classifier in IAG-Net. During testing, the predicted image-level label $\hat{\mathbbm{y}}$ and the predicted instance-level label $\hat{y}_\mathbf{x}$ under both of the two ablations are also obtained by Eq.~\ref{eq:image_pred} and Eq.~\ref{eq:voxel_pred}, respectively.

Results with VGG-Net are shown in Table~\ref{Tab:ablation1}. We observe that {jointly training the global and the local classifiers achieves better PDAC segmentation results}. Training with only the global classifier leads to comparable classification results to jointly training ({\em{i.e.}}, Sensitivity: 98.00\%, Specificity: 98.75\% vs. Sensitivity: 99.25\%, Specificity: 97.50\%). But surprisingly, it does not work for PDAC segmentation (instance probabilities are uniformly distributed over all locations). \textbf{The reason might be that the existence of PDAC makes the whole pancreas abnormal.} 
Without the local classifier, {the} global classifier alone cannot segment PDAC regions. Compared with jointly training, training with only local classifiers is confronted with a slight performance drop (-3.19\%) for PDAC segmentation, and a significant specificity drop (-68\%) for normal/PDAC classification. This is reasonable, since local classifiers aim at localizing the PDAC masses, which is more prone to identify a case as a PDAC case. These observations verify the importance of our joint learning for PDAC prediction\vspace{0.6em}.

\begin{table}[t]
\setlength{\tabcolsep}{5pt}
\small
\centering
\caption{{{Ablation on the joint learning framework.}}}
\label{Tab:ablation1}
\resizebox{1\linewidth}{!}{
\begin{tabular}{ccccc}
\toprule[0.15em]
Global & Local & \multirow{2}{*}{Mean DSC (Max, Median)} & \multirow{2}{*}{Sens.} & \multirow{2}{*}{Spec.}\\
classifier & classifier & & & \\
\midrule
 & \checkmark & 51.19 $\pm$ 18.49 (86.80, 52.95) & 98.00 & 29.50 \\
\checkmark & & 6.900 $\pm$ 6.100 (47.04, 5.210) & 98.00 & 98.75 \\
\checkmark & \checkmark & {54.38}~$\pm$~{18.77} (87.65, 57.00) & 99.25 & 97.50 \\
\bottomrule[0.15em]
\end{tabular}
}
\end{table}

\noindent\textbf{(2) Ablation study on the global classifier.}

To show the effectiveness of our global classifier by attention-guided MIL for PDAC segmentation, we replace it with max-pooling and average-pooling MIL. More specifically, after acquiring the attention $a_\mathbf{x}$ for instance $\mathbf{x}$, on one hand, $a_\mathbf{x}$ can be supervised by per-voxel annotation $y_\mathbf{x}$, if any. On the other hand, $a_\mathbf{x}$ is fed into a {max/average} pooling operation.
The probability that an image $\mathbf{Z}$ contains PDAC masses is 
\begin{equation}
    P_g(\mathbf{Z};\bm{\Theta},\mathbf{w}_a) = \frac{1}{1+\exp{(-\mathbf{g}(\mathbf{Z};\bm{\Theta},\mathbf{w}_a))}},
\end{equation}
where $\mathbf{g}(\mathbf{Z};\bm{\Theta},\mathbf{w}_a) = \max_{\mathbf{x}\in\mathcal{X}}a_\mathbf{x}(\mathbf{Z};\bm{\Theta},\mathbf{w}_a)$ or $\mathbf{g}(\mathbf{Z};\bm{\Theta},\mathbf{w}_a) = \frac{\Sigma_{\mathbf{x}\in\mathcal{X}}a_\mathbf{x}(\mathbf{Z};\bm{\Theta},\mathbf{w}_a)}{|\mathcal{X}|}$. Results with VGG-Net are shown in Table~\ref{Tab:ablation2}. Attention-guided MIL achieves better prediction results than max/average-pooling MIL\vspace{0.6em}.

\begin{table}[t]
\renewcommand\arraystretch{1.1}
\small
\centering
\caption{{{Ablation study on the global classifier.}}}
\label{Tab:ablation2}
\begin{tabular}{lccc}
\toprule[0.15em]
MIL method & Mean DSC (Max, Median) & Sens. & Spec.\\
\midrule
Max & ~51.85 $\pm$ 20.63 (87.65, 55.21)& 98.75 & 96.50\\
Average &~34.12 $\pm$ 19.46 (83.25, 32.91) & 86.75 & 99.75\\
Attention & ~{54.38}~$\pm$~{18.77} (87.65, 57.00) & 99.25 & 97.50\\
\bottomrule[0.15em]
\end{tabular}
\end{table}

\noindent\textbf{(3) Ablation study on the local classifier.}

\noindent\textbf{(\romannumeral1) What if we train it on per-voxel pseudo-labels?}
To train the local classifier on per-voxel pseudo-labels, we follow the typical teacher-student model~\cite{Lee2013pseudo} for self-training. We remove the second stream of IAG-Net and train the explicitly-learned attention on the training data with per-voxel annotations as the teacher model. Then, we use the teacher model to generate per-voxel pseudo-labels for the training data without per-voxel annotation. After that, We re-train the explicitly-learned attention by including the generated per-voxel pseudo-labels as the student model. The first stream is still supervised by image-level annotations when training both teacher and student models. {This method is termed as IAG-PVPL-Net (PVPL: Per-Voxel Pseudo-Labels)}. The classification and segmentation results are given by $P_g$ and $p_\mathbf{x}$, respectively. Training the local classifier on per-voxel pseudo-labels obtains 51.74 $\pm$ 20.40 (87.16, 54.96) for mean (\% max, median) DSC, 98.75\% sensitivity and 97.25\% specificity, leading to around 3\% DSC drop compared with IAG-Net, while it requires much more training iterations.

\begin{figure}[t]
\begin{center}
    \includegraphics[width=1\linewidth]{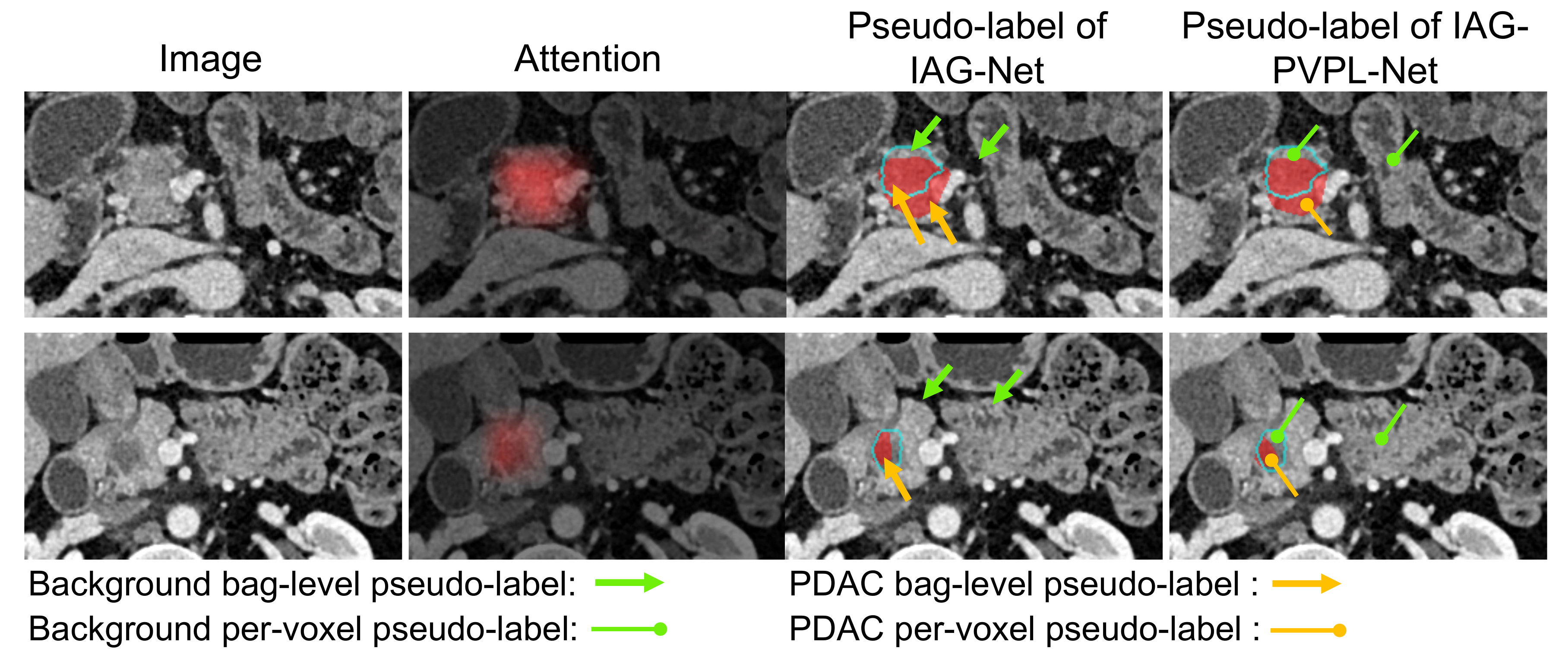}
\end{center}
\caption{{Two examples of the attention and pseudo-labels of unlabeled training cases. Each row shows {an input} image, attention values of IAG-Net, bag-level pseudo-label of IAG-Net and per-voxel pseudo-label of IAG-PVPL-Net. The blue contours are the boundaries of the annotated PDAC segmentation masks (shown as references, which are not used during training). Since usually the misclassified instances in a bag are minor, the correctness of the bag-level pseudo-labels can be guaranteed after average pooling. But the per-voxel pseudo-labels of these instances are treated as noises for IAG-PVPL-Net.}} 
\label{Fig:Pseudolabel}
\end{figure}

{To better understand why the bag-level pseudo-labels are better than the per-voxel pseudo-labels, some of the pseudo-labels generated by IAG-Net and IAG-PVPL-Net are shown in Fig.~6. Bag-level pseudo-labels obtained by average pooling impose soft constraints on instances whose scores are low. For example, among 100 instances, 95 instances are scored $1.0$, and the rest 5 are scored $0.0$. Then the loss w.r.t a bag-level pseudo-label is $-\log(0.95)$. But per-voxel pseudo-labels impose strict constraints on each instance, \emph{i.e.}, the losses of the instances with very low scores w.r.t. per-voxel pseudo-labels are nearly $\infty$. Consequently, the bag-level pseudo-labels are better for training\vspace{0.6em}.}

\noindent\textbf{(\romannumeral2) Is the attention-based weight factor $\lambda$ better than a constant weight factor?}
We set the weight factor $\lambda$ to a default constant: $\lambda=1.0$. The result is: 52.69$\pm$ 18.11 Mean, 86.39 max and 54.18 median DSC (\%) for PDAC segmentation, and 98.5\% sensitivity and 97.75\% specificity for normal/PDAC classification. Compared with the attention-based weight factor, it leads to 1.69\% mean DSC drop\vspace{0.6em}.

\noindent\textbf{(\romannumeral3) What if we do not train the local classifier on data without per-voxel annotations?}
We set $|\mathcal{D}^u|=0$ during training. The PDAC segmentation results based on both VGG-Net and U-Net are shown in Table~\ref{Tab:ablation3-1}. Training the local classifier on $\mathcal{D}^u$ can always boost the PDAC segmentation performance (+4.55\% for VGG-Net and +5.24\% for U-Net)\vspace{0.6em}.

\begin{table}[t]
\renewcommand\arraystretch{1.1}
\centering
\caption{{{Ablation on the local classifier.}}}
\label{Tab:ablation3-1}
\resizebox{1\linewidth}{!}{
\begin{tabular}{lllccc}
\toprule[0.15em]
Backbone & $|\mathcal{D}^l|$ & $|\mathcal{D}^u|$ & Mean DSC (Max, Median) & Sens. & Spec.\\
\midrule[0.1em]
\multirow{2}{*}{VGG-Net} & 150 & 0 & 49.83 $\pm$ 20.59 (86.80, 52.77)& 98.75 & 98.00 \\
 & 150 & 150 & 54.38~$\pm$~18.77 (87.65, 57.00) & 99.25 & 97.50 \\
\midrule
\multirow{2}{*}{U-Net} & 150 & 0 & 55.05 $\pm$ 25.13 (94.15, 59.81) & 98.50& 97.25 \\
 & 150 & 150 & 60.29 $\pm$ 21.60 (94.04, 64.37) & 99.75 & 96.50 \\
\bottomrule[0.15em]
\end{tabular}
}
\end{table}

\noindent\textbf{(\romannumeral4) How does the PDAC segmentation performance change by varying the ratio between $|\mathcal{D}^l|$ and $|\mathcal{D}^u|$?}
Since each fold of our PDAC dataset has 300 PDAC cases, $|\mathcal{D}^l| + |\mathcal{D}^u| \leq 300$. We train our IAG-Net by varying the ratio between $|\mathcal{D}^l|$ and $|\mathcal{D}^u|$. Table~\ref{Tab:ablation3-2} shows the PDAC segmentation performance of our IAG-Net is improved as $|\mathcal{D}^l|$ is increased, which is not surprising. But we also observe that training the local classifier on $\mathcal{D}^u$ always significantly boosts the PDAC segmentation performance, which is even approaching the result obtained by fully-supervised segmentation (the last row)\vspace{0.6em}.

\begin{table}[t]
\setlength{\tabcolsep}{5pt}
\small
\centering
\caption{{{Ablation on the local classifier by varying $|\mathcal{D}^l|$ and $|\mathcal{D}^u|$ for each fold with VGG-Net}}.}
\label{Tab:ablation3-2}
\begin{tabular}{llcccc}
\toprule[0.15em]
$|\mathcal{D}^l|$ & $|\mathcal{D}^u|$ & Mean DSC (Max, Median) & Sens. & Spec.\\
\midrule[0.1em]
50 & 0 & 44.31 $\pm$ 21.29 (82.37, 46.46) & 98.25 & 97.75 \\
50 & 250 & 49.54 $\pm$ 22.87 (85.63, 52.54) & 98.50 & 97.25 \\
\midrule
150 & 0 & 49.83 $\pm$ 20.59 (86.80, 52.77)& 98.75 & 98.00 \\
150 & 150 & 54.38~$\pm$~18.77 (87.65, 57.00) & 99.25 & 97.50 \\
\midrule
300 & 0 & 55.45 $\pm$ 19.60 (88.36, 58.29) & 99.25 & 97.75\\
\bottomrule[0.15em]
\end{tabular}
\end{table}

\noindent\textbf{(\romannumeral5) Does the attention guidance ($p_\mathbf{x}$) itself have a good segmentation ability in IAG-Net?}
Using $p_\mathbf{x}$ or $q_\mathbf{x}$ alone in Eq.~\ref{eq:voxel_pred} leads to $54.28 \pm 19.04$ or $54.38 \pm 18.53$ mean DSC (\%), which is comparable to the result by combining $p_\mathbf{x}$ and $q_\mathbf{x}$ (Eq.~\ref{eq:voxel_pred}). This shows that the attention guidance itself has a good segmentation ability.

\subsection{{Discussions}}
{For the early stage of a PDAC case, as the PDAC mass is subtle, obtaining an accurate contour for PDAC segmentation is difficult. But, the pancreas may have some abnormal changes, \emph{e.g.}, duct dilation or texture abnormality. PDAC segmentation is a much more challenging task than normal/PDAC classification. This may be the reason that although PDAC segmentation results are not very high by using even state-of-the-art methods, we can still achieve good classification results.}

{Our IAG-Net can be applied to multi-class segmentation, \emph{i.e.}, to segment the entire pancreas and PDAC simultaneously. Let us define an attention map as a matrix consisting of the attention values of all the instances on the feature map $\mathbf{F}(\mathbf{Z};\bm\Theta)$. The key to the extension is to explicitly learn two attention maps under the supervision of the per-voxel annotations for both the pancreas and the PDAC, respectively. How to explicitly learn an attention map is shown in Sec. \ref{Sec:el_attention}. Then the global classifier which targets normal/PDAC classification, is also trained based on the attention map of the PDAC, as the same as Sec. \ref{sec:attention_MIL}. For the local classifier, we train two ``one-vs-all'' classifiers for semi-supervised pancreas segmentation and semi-supervised PDAC segmentation, based on the attention maps of the pancreas and the PDAC, respectively. The training process of each of the two ``one-vs-all'' classifiers is the same as the training process shown in Sec. \ref{sec:local_classifier}. The loss function for the local classifier is the sum of the losses of the two ``one-vs-all'' classifiers.}

\begin{figure}[t]
\begin{center}
    \includegraphics[width=0.9\linewidth]{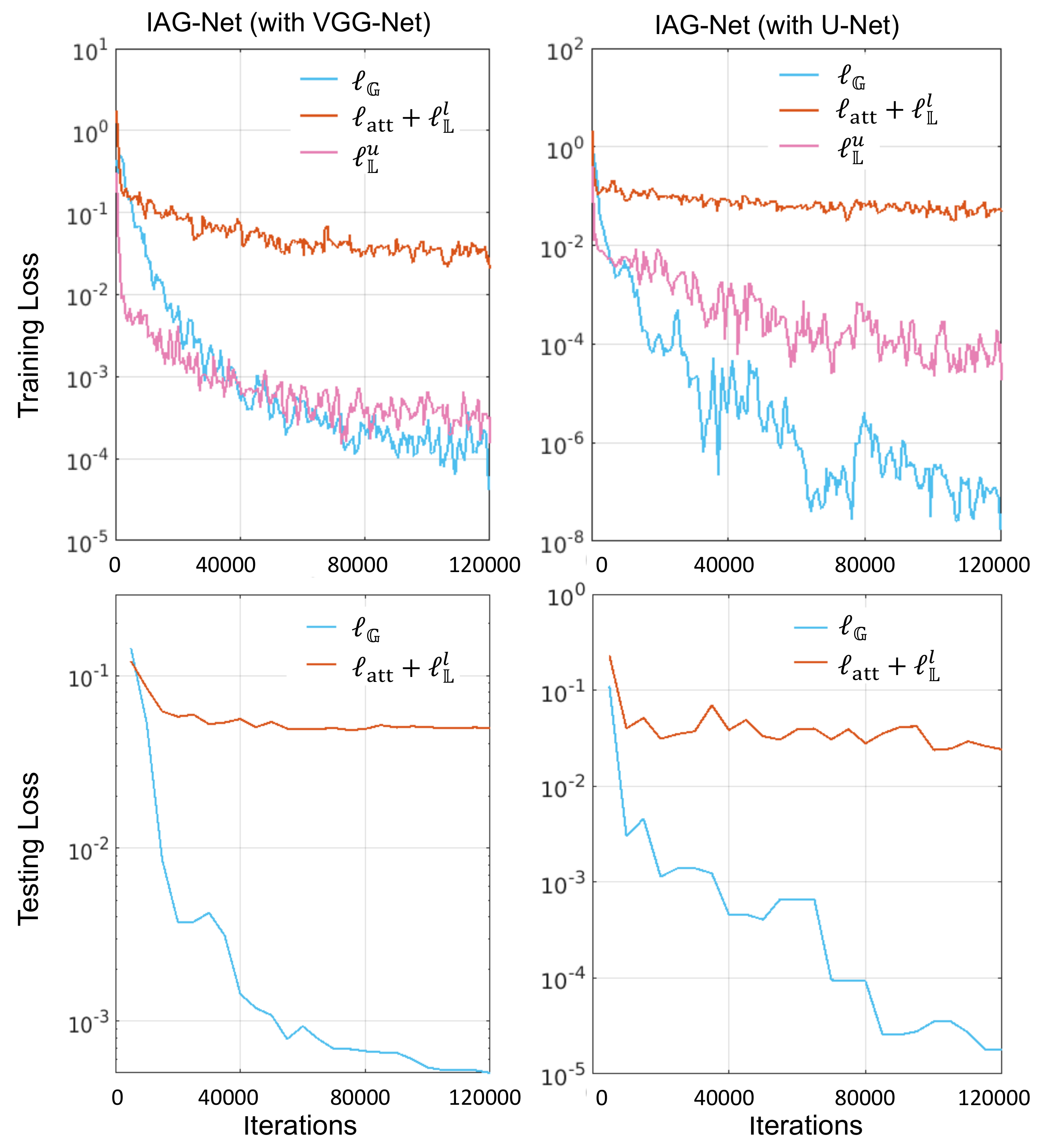}
\end{center}
\vspace{-0.5em}
\caption{Training {and testing loss} curves of IAG-Net with VGG-Net and U-Net as backbones.} 
\label{Fig:Curve}
\end{figure}

{Finally, Fig.~\ref{Fig:Curve} shows training {and testing} curves of the image-level classification loss $\ell_\mathbbm{G}$ (Eq.~\ref{eq:mil_loss}), instance-level segmentation loss for training data with per-voxel annotations $\ell_{att}$ + $\ell^l_\mathbbm{L}$ (Eq.~\ref{eq:att_loss} + Eq.~\ref{eq:SSL_labeled}), and the instance-level segmentation loss for training} {data without per-voxel annotations $\ell^u_\mathbbm{L}$ (Eq.~\ref{eq:SSL_unlabeled}). {Training losses are obtained via mini-batches every 20 iterations, and testing losses are acquired from the whole testing set every 5000 iterations. There is no $\ell^u_\mathbbm{L}$ loss for testing.} We observe that all losses converge after a certain number of iterations{, and the testing losses decrease accordingly with the decrease of the training losses.} The Y-axis in the plots are shown in the log space, so the loss values are very small when approaching 120,000 iterations.}

\subsection{IAG-Net on Semi-supervised Tumor Segmentation}
To verify the generality of IAG-Net on semi-supervised pancreatic tumor segmentation, we test it on the pancreas tumor segmentation dataset in MSD challenge \cite{Simpson19a}. There are two targets in the pancreas tumor dataset: pancreas and tumor. Here we only focus on tumor segmentation, which consists of multiple types of pancreatic tumors. 282 labeled data are released for training and validation. Noted that in this dataset, there are no normal cases, thus we exclude the global classifier in our IAG-Net for semi-supervised tumor segmentation only.

We randomly partition the dataset into 4 folds. The numbers of the cases belonging to each fold are 70, 70, 71 and 71. Following the same setting as we used for the JHMI dataset, we randomly choose half of the training data as the data without per-voxel annotation during training, which leaves only 35 cases with per-voxel annotation for each fold. The 2D U-Net is adopted as the backbone network unless otherwise specified. We train and test on the axial view of each case. {We compare our IAG-Net with other competitors, and the results are shown in Table~\ref{Tab:MSD_comparebase}. IAG-Net outperforms other methods by a large margin.} {IAG-Net (fully-supervised) in Table~\ref{Tab:MSD_comparebase} is also shown as a reference for comparison.}

\begin{table}[t!]
\renewcommand\arraystretch{0.95}
\small
\centering
\caption{{Comparison on pancreas tumor dataset in MSD challenge.}}
\label{Tab:MSD_comparebase}
\resizebox{1\linewidth}{!}{
\begin{tabular}{lc}
\toprule[0.15em]
{Method} & {Mean DSC (Max, Median)} \\
\midrule
{Li \emph{et al.} \cite{Li2018thoracic}} & {19.04~$\pm$~24.48 (85.06, 3.17)} \\
{DMPCT \cite{Zhou2019semi}} & {25.60~$\pm$~25.20 (89.32, 23.75)} \\
{Collaborative \cite{Zhou2019collaborative}} & {22.42~$\pm$~24.86 (87.67, 13.17)} \\
{Consistency-based \cite{Ouali2020semi}} & {24.63~$\pm$~27.10 (90.37, 15.12)}\\
{Consistency-self-ensembling \cite{Ref:LiYCFH20}} & {30.01~$\pm$~27.55 (90.07, 28.40)}\\
{IAG-Net (Ours) }& {32.49~$\pm$~27.82 (94.08, 31.41)}\\
\cmidrule(lr){1-2}
{IAG-Net (fully-supervised)} & {33.91~$\pm$~28.21 (90.83, 34.25)} \\
\bottomrule[0.15em]
\end{tabular}
}
\vspace{1em}
\end{table} 

{We also vary the ratio between $|\mathcal{D}^l|$ and $|\mathcal{D}^u|$}. Results are summarized in Table~\ref{Tab:MSD}, which show that our IAG-Net can improve the tumor segmentation result by leveraging the training data without per-voxel annotations (around 2\% accuracy gain) and even surpasses the fully-supervised method in terms of max DSC. 

\begin{table}[t!]
\renewcommand\arraystretch{0.95}
\small
\centering
\caption{Comparison on pancreas tumor dataset in MSD challenge {by varying $|\mathcal{D}^l|$ and $|\mathcal{D}^u|$}. Each fold has 70 or 71 cases. $|\mathcal{D}^l| + |\mathcal{D}^u| \leq$ 70/71.}
\label{Tab:MSD}
\begin{tabular}{llc}
\toprule[0.15em]
$|\mathcal{D}^l|$ & $|\mathcal{D}^u|$ & {Mean DSC (Max, Median)} \\
\midrule
35 & 0 & 30.65~$\pm$~27.44 (92.37, 29.60) \\
35 & 35/36 & 32.49~$\pm$~27.82 (94.08, 31.41) \\
70/71 & 0 & 33.91~$\pm$~28.21 (90.83, 34.25) \\
\bottomrule[0.15em]
\end{tabular}
\vspace{1em}
\end{table}

\section{Conclusions}
\label{Conclusions}
This paper addresses the problem of PDAC prediction \emph{i.e.}, normal/PDAC classification and PDAC segmentation under the partially supervised setting.
We {present} an Inductive Attention Guidance (IAG) strategy for learning a global image-level classifier for normal/PDAC segmentation and a local instance-level classifier for semi-supervised PDAC segmentation, which enjoys the advantages of bridging the MIL-based global and local classifiers. We showed empirically on {the} JHMI dataset the superiority of the proposed IAG-Net for PDAC prediction, which is helpful to computer-assisted clinical diagnoses. Additionally, we verified the generality of IAG-Net on {the} pancreas tumor segmentation dataset in MSD challenge.



\bibliographystyle{IEEEtran}

\bibliography{bare_jrnl}

\end{document}